\documentclass[11pt]{article}

\usepackage{epsfig,amsmath,latexsym,amssymb}
\usepackage{graphicx}
\usepackage{lscape}
\usepackage{picture, eso-pic, tikz} 
\usepackage{dsfont}
\usepackage{listings}
\usepackage{changes}
\usepackage{hyperref}
\usepackage{bbding}

\renewcommand{\baselinestretch}{1.1}
\def\R{{\mathbb R}}  
\def\N{{\mathbb N}}  
\def\E{{\mathbb E}}  %

\newcommand{\Remm}[1]{}
\newtheorem{theo}{Theorem}[section]

\newtheorem{model ass}[theo]{Model Assumptions}

\newtheorem{example}[theo]{Example}

\newtheorem{rems}[theo]{Remarks}

\def\EndExample{\hfill {\scriptsize $\blacksquare$}}
\numberwithin{equation}{section}

\definecolor{MyGray}{rgb}{0.92,0.92,0.92}


\definecolor{British racing}{rgb}{0.0, 0.5, 0.0}
\def\bx{\boldsymbol{x}}
\def\be{\boldsymbol{e}}

\def\bX{\boldsymbol{X}}

\def\b0{\boldsymbol{0}}

\def\bbb{\boldsymbol{b}}

\def\bvartheta{\boldsymbol{\vartheta}}

\def\bbeta{\boldsymbol{\beta}}

\def\bnu{\boldsymbol{\nu}}

\def\b0{\boldsymbol{0}}

\def\bm{\boldsymbol{m}}

\usepackage{todonotes}
\newcommand{\Comments}{1}
\newcommand{\mynote}[2]{\ifnum\Comments=1\textcolor{#1}{#2}\fi}
\newcommand{\mytodo}[2]{\ifnum\Comments=1%
  \todo[linecolor=#1!80!black,backgroundcolor=#1,bordercolor=#1!80!black]{#2}\fi}

\ifnum\Comments=1               
\fi

\ifnum\Comments=1               
\fi

\ifnum\Comments=1               
\fi

\begin{document}
\author{Ronald Richman\footnote{Old Mutual Insure and University of the Witwatersrand, Johannesburg, South Africa; ronaldrichman@gmail.com}
	\and
  Mario V.~W\"uthrich\footnote{RiskLab, Department of Mathematics, ETH Zurich,
mario.wuethrich@math.ethz.ch}}

\date{Version of \today}
\title{Conditional Expectation Network for SHAP}
\maketitle

\begin{abstract}
\noindent  
A very popular model-agnostic technique for explaining predictive
models is the {\bf SH}apley {\bf A}dditive ex{\bf P}lanation (SHAP).
The two most popular versions of SHAP are a conditional expectation version
and an unconditional expectation version (the latter is also known as
interventional SHAP). Except for tree-based methods, usually 
the unconditional
version is used (for computational reasons). We provide a 
(surrogate) neural
network approach which allows us to efficiently calculate the
conditional version for both neural networks and other regression
models, and which properly considers the dependence
structure in the feature components. This proposal is also useful to provide
{\tt drop1} and {\tt anova} analyses in complex regression
models which are similar to their generalized linear model
(GLM) counterparts, and we provide a partial dependence plot (PDP)
counterpart that considers the right dependence structure in the
feature components.

~

\end{abstract}

\section{Introduction}
We start by formally introducing the problem studied in this paper.
We consider a random tuple $(Y,\bX)$ with $Y$ describing
a real-valued response variable that is supported by features $\bX=(X_1,\ldots, X_q)^\top$
taking values in a feature
space ${\cal X} \subseteq \R^q$. We use
these features to build a regression model
\begin{equation}\label{general regression function}
\mu: {\cal X} \to \R, \qquad 
\bx \mapsto \mu(\bx),
\end{equation}
that serves at predicting the response variable $Y$, e.g., we can use the conditional
expectation regression function defined by
\begin{equation*}
\bx~\mapsto ~
\mu(\bx) = \E \left[ \left. Y \right| \bX= \bx\right].
\end{equation*} 
The problem that we solve in this paper is the 
computation of the conditionally expected regression function \eqref{general regression function}, 
if only a subset
of the feature components of $\bX=(X_1,\ldots, X_q)^\top$ is available.
E.g., if only the first two components $(X_1,X_2)=(x_1,x_2)$ of $\bX$ are observed, 
we would like to compute the conditional expectation
\begin{equation}\label{conditional expectations X2}
\E \left[ \left. \mu(\bX) \right| X_1=x_1, X_2=x_2\right],
\end{equation}
this is further motivated in Example \ref{mean regression function}, below.
Such conditional expectations \eqref{conditional expectations X2} are
of interest in many practical problems, e.g., they enter the
{\bf SH}apley {\bf A}dditive ex{\bf P}lanation (SHAP) of Lundberg--Lee \cite{lundberg2017}, see also Aas et al.~\cite{Aas},
they are of interest in discrimination-free insurance pricing, see Lindholm et al.~\cite{Lindholm1},
and they are also useful in a variable importance analysis, similar to the
{\tt anova} (analysis-of-variance/analysis-of-deviance) and the {\tt drop1} analyses for generalized linear models (GLMs)
in the {\sf R} statistical software \cite{Cran}, see also Section 2.3.2
in McCullagh--Nelder \cite{MN}. Moreover, it is known that the partial dependence plot (PDP) of
  Friedman \cite{friedman2001greedy} and Zhao--Hastie \cite{zhao2019causal} for marginal explanation cannot
  correctly reflect the dependence structure in the features $\bX$. Below, we provide an alternative proposal,
  called marginal conditional expectation plot (MCEP), that mitigates this deficiency.

The main difficulty in efficiently evaluating \eqref{conditional expectations X2} is that
it has a similar computational complexity as nested simulations, if one wants to calculate these conditional expectations
\eqref{conditional expectations X2} with Monte Carlo simulations for all possible
values $(x_1,x_2)$. That is, we need to simulate for {\it all} values $(x_1,x_2)$ the random vector
$(X_3,\ldots, X_q)$ conditionally, given $(X_1,X_2)=(x_1,x_2)$.
Such nested simulations are computationally expensive, and things become even
more involved if we need to calculate these conditional expectations for {\it all} possible
subsets of the components of $\bX$, e.g., when performing
a SHAP analysis. A further difficulty is that these simulations
can only be done by bootstrapping if the true probability law $\pi$ of $\bX$ is unknown, i.e., 
if we have to rely on an i.i.d.~sample $(\bX_i)_{i=1}^n$ of $\bX \sim \pi$. In that case sparsity of
observations in different parts of the feature space ${\cal X}$ poses another
issue, and this issue becomes more serious for higher dimensional features ${\cal X}$.

In financial mathematics, a similar problem occurs when evaluating American options.
Carriere \cite{Carriere}, Longstaff--Schwartz \cite{Longstaff} and 
Tsitsiklis--Van Roy \cite{Tsitsiklis} have proposed to map the (inner) nested valuation
problem to a suitable class of basis functions; this is also known as least squares Monte Carlo (LSMC).
Basically, this means that \eqref{conditional expectations X2}
should be expressed as a new regression function in the variables $(x_1,x_2)$, and this
new (inner) regression function still needs to be determined.
Benefiting from the huge modeling flexibility of neural networks (universal approximation property), it has been proposed to
use a neural network as this new inner regression function; see, e.g., Cheridito et al.~\cite{Cheridito},
Krah et al.~\cite{Krah} or Jonen et al.~\cite{Jonen}. These proposals have been made for a fixed set of observable
components of $\bX$. 

Our main contribution extends the set-up of
Cheridito et al.~\cite{Cheridito} to simultaneously
model the conditional expectations of all possible subsets of observable components of $\bX$. This allows us to develop a fast algorithm for
estimating conditional SHAP, based on a surrogate neural network; this
proposal works both for neural networks and other predictive machine learning
algorithms. 
We discuss the necessary variable masking used, and we propose a specific fitting procedure
so that the extreme cases about the full knowledge of $\bX$ and about the null model (not knowing $\bX$) are correctly
calibrated. Furthermore, we present applications of this approach to model-agnostic variable
importance tools, such as an {\tt anova} or a {\tt drop1} analysis, similar to
GLMs, MCEPs similar to PDPs,  and a global conditional SHAP decomposition of the generalization
loss.

\medskip

{\bf Organization.}
In the next section, we introduce the surrogate neural network
for conditional expectation estimation, and we discuss the specific fitting
procedure so that the extreme cases of full knowledge and of zero knowledge
are properly calibrated. In Section \ref{Example: variable importance},
we apply these conditional expectations to analyses variable importance
in predictive models. For this we introduce an {\tt anova} analysis and
a {\tt drop1} analysis that are similar to their GLM counterparts, see Section \ref{anova and drop1 analyses}.
In Section \ref{sec: MCEP}, we introduce the marginal conditional expectation plot (MCEP) as a conditional
expectation counterpart of the partial dependence plot (PDP) that properly considers the dependence
structure in the features.
In Section \ref{Example: SHAP}, we discuss the application of our
proposal to efficiently calculate the conditional SHAP. On the one hand,
we consider the individual (local) mean decomposition, and on the
other hand a global fair SHAP score decomposition for variable
importance. Finally, Section \ref{Conclusions} concludes.

\section{Conditional expectation network}
We begin from a given regression function $\bx \mapsto \mu(\bx)$,
which maps the feature values $\bx=(x_1,\ldots, x_q)^\top \in {\cal X} \subseteq \R^q$ to real-valued output values $\mu(\bx) \in \R$; we assume that this
regression function $\mu$ is given. In our example below, this regression function
has been constructed within a fixed (given) network architecture based
on an i.i.d.~sample $(Y_i,\bx_i)_{i=1}^n$. However, this is not
an essential point in our proposal. The following methodology can be applied
to any other regression model, such as gradient boosting trees,
nonetheless, if the regression function
$\bx \mapsto \mu(\bx)$ comes from a neural network, it will
speed up the following (network) model fitting procedure, because the 
gradient descent algorithm can be initialized with exactly the network
weights as used in the (first) regression function $\mu$.

Assume that $\bX \sim \pi$ denotes the random selection of
a feature value $\bX=\bx \in {\cal X}$.
Select a subset ${\cal C} \subseteq {\cal Q}:= \{1,\ldots, q\}$ of the feature
component indices. We generically write $\bX_{\cal C}=(X_j)_{j \in {\cal C}}$ for selecting
the components of $\bX$ with indices $j\in {\cal C}$.
Our goal is to calculate the conditional
expectations
\begin{equation}\label{conditional expectation}
\mu_{\cal C}(\bx):=
\E \left[ \left.\mu(\bX) \right| \bX_{\cal C}=\bx_{\cal C} \right],
\end{equation}
with the two extreme cases ${\cal C}=\emptyset$ and ${\cal C}={\cal Q}$
given by
\begin{equation}\label{true and null model}
\mu_0 := \mu_{\emptyset}(\bx)=\E[\mu(\bX)]\qquad \text{ and } \qquad
\mu(\bx)=\mu_{\cal Q}(\bx)=\E \left[\left. \mu(\bX) \right| \bX=\bx \right].
\end{equation}
The former is called the {\it null model} and the latter is the {\it full model}.
In general, the conditional expectation $\mu_{\cal C}(\bx)$ 
cannot easily be calculated because in regression function
$\bx\mapsto \mu(\bx)$, we cannot simply ``turn off'' the components
of $\bx$ that are not in ${\cal C}$.

\begin{example}\label{mean regression function}\normalfont
Assume $Y$ is an integrable random variable and that the full model is given by the conditional expectation regression function
\begin{equation}\label{conditional expectation regression example}
\bx~\mapsto ~
\mu(\bx) = \E \left[ \left. Y \right| \bX= \bx\right].
\end{equation}   
If, for some reason, only the components $\bX_{\cal C}$ of $\bX$, ${\cal C} \subset {\cal Q}$,  have been observed,
then we can only build a regression model
\begin{equation*}
\bx_{\cal C}= (x_j)_{j \in {\cal C}}~\mapsto~
\E \left[\left. Y \right| \bX_{\cal C}=\bx_{\cal C} \right] =
\E \left[\left. \mu(\bX) \right| \bX_{\cal C}=\bx_{\cal C} \right]=\mu_{\cal C}(\bx),
\end{equation*}
where we have used the tower property of conditional expectations. Thus, the conditional expectation
\eqref{conditional expectation} naturally arises under partial information.
Note that the full model $\mu=\mu_{\cal Q}$ given in \eqref{conditional expectation regression example}
dominates in convex order any other regression function
$\mu_{\cal C}$, ${\cal C} \subset {\cal Q}$, i.e., it has a higher resolution than any other conditional expectation
regression function; see also Theorem 2.27 in Gneiting--Resin \cite{GneitingResin}
for the resolution (discrimination) of a regression model.
\EndExample
\end{example}

We assume that all considered random variables are square integrable.
This implies that we can work on a Hilbert space. We then
receive the conditional expectation $\mu_{\cal C}(\bX)$
as the orthogonal projection of $\mu(\bX)$ onto the
subspace $\sigma(\mu_{\cal C}(\bX))$
generated by random variable $\mu_{\cal C}(\bX)$ in this Hilbert space.
That is, the conditional expectation is the measurable
function $\bx_{\cal C}=(x_j)_{j \in {\cal C}} \mapsto \mu_{\cal C}(\bx)$
that minimizes the mean squared distance
\begin{equation*}
\E \left[ \left(\mu(\bX)- \mu_{\cal C}(\bX)\right)^2 \right]
~\stackrel{\rm !}{=}~\min.
\end{equation*}
Among all $\bx_{\cal C}$-measurable functions, this
conditional expectation is obtained by
the solution of
\begin{equation}\label{true square loss}
\mu_{\cal C}(\bx) ~=~
\underset{\widehat{\mu}}{\arg\min}~
\E \left[\left. \left(\mu(\bX) - \widehat{\mu}\right)^2
\right| \bX_{\cal C}=\bx_{\cal C}\right],
\end{equation}
for $\pi$-a.e.~$\bx_{\cal C}$.
The idea now is to approximate the functions
$\bx_{\cal C} \mapsto \mu_{\cal C}(\bx)$
simultaneously for all subsets ${\cal C}\subseteq {\cal Q}$
by a neural network 
\begin{equation}\label{neural network}
\bx ~\mapsto ~ {\rm NN}_{\bvartheta}(\bx),
\end{equation}
where ${\rm NN}_{\bvartheta}$ denotes a neural network of
a fixed architecture with network weights (parameter) $\bvartheta$.
There are two important points to be discussed:
\begin{itemize}
\item[(1)] The neural network \eqref{neural network} considers
all components of input $\bx$. In order that this network can 
approximate $\mu_{\cal C}(\bx)$, we need to mask all
components in $\bx=(x_1,\ldots, x_q)^\top$ which are not contained in ${\cal C}$.
Choose a mask value $\bm=(m_1,\ldots, m_q)^\top \in \R^q$, the specific choice is going to
be discussed below; for the moment, it should just
be a sort of ``neutral value''.
We set
\begin{equation}\label{masked x}
\bx^{(\bm)}_{\cal C} := \left(m_1 + (x_1-m_1) \mathds{1}_{\{1 \in {\cal C}\}}, 
\ldots, m_q+(x_q-m_q) \mathds{1}_{\{q \in {\cal C}\}}\right)^\top.
\end{equation}
We then try to find an optimal network parameter $\bvartheta$
such that ${\rm NN}_{\bvartheta}(\bx^{(\bm)}_{\cal C})$ is a good approximation
to $\mu_{\cal C}(\bx)$ for all features $\bx \in {\cal X}$ and all
subsets ${\cal C} \subseteq {\cal Q}$.

\item[(2)] In our applications, we explore an empirical version of \eqref{true square loss}
  to find the optimal network parameter $\bvartheta$. Assume we have observed features $(\bx_i)_{i=1}^n$.
  Then, we solve
\begin{equation}\label{optimal network parameter}
\widehat{\bvartheta}~=~\underset{\bvartheta}{\arg\min}~
\frac{1}{3n}\,
\sum_{l=1}^{3n} \Big(\widetilde{\mu}(\bx^{[3]}_l) -   
{\rm NN}_{\bvartheta}(\bx^{[3]}_l)\Big)^2,
\end{equation}
where, to both calibrate the network and meet the logical constraints of the
full and the null model, we triplicate the observed features $(\bx_i)_{i=1}^n$ as 
follows:
\begin{itemize}
\item[(a)] For $1\le l \le n$, we set $\bx^{[3]}_{l}=\bx_l$
and $\widetilde{\mu}(\bx^{[3]}_l)=\mu(\bx_l)$. These
instances are used to ensure that we can replicate the
full model, see \eqref{true and null model}. 
\item[(b)] For $n+1\le l \le 2n$, we set $\bx^{[3]}_{l}=\bm$ and $\widetilde{\mu}(\bx^{[3]}_l)=\mu_0$. These
instances are used to ensure that we can replicate the
null model, see \eqref{true and null model}. If $\mu_0$ is not available we just take the empirical
mean of $(\mu(\bx_i))_{i=1}^n$ as its estimate.
\item[(c)] For $2n+1\le l \le 3n$, we set $\bx^{[3]}_{l}=\bx^{(\bm)}_{l-2n,{\cal C}_l}$,
  see \eqref{masked x}, and $\widetilde{\mu}(\bx^{[3]}_l)=\mu(\bx_l)$,
where the sets ${\cal C}_l \subseteq {\cal Q}$ are chosen randomly and
independently such that they mask independently of all other
components each component $j\in {\cal Q}$ with probability $1/2$.
\end{itemize}
\end{itemize}

\begin{rems}\normalfont
\begin{itemize}
\item We use the above cases (a) and (b) with indices $1\le l \le 2n$
to ensure that the extreme cases \eqref{true and null model}, the
null model $\mu_0$ and the full model $\mu(\bx)$, can
be approximated by the estimated neural network ${\rm NN}_{\widehat{\bvartheta}}$.
These two cases serve as calibration of the conditional expectations.
\item The above case (c) with indices $2n+1\le l \le 3n$ models the conditional expectations \eqref{conditional expectation}, where
the input data $\bx^{[3]}_{l}=\bx^{(\bm)}_{l-2n,{\cal C}_l}$ has randomly masked components $m_j$ with $j\not\in {\cal C}_l$, and it tries to 
approximate the full model as well as possible.
\item In relation to the previous item, there is some connection
to masked auto-encoders which randomly mask part of the input images,
which are then reconstructed by the auto-encoders; see He et al.~\cite{He}.
These masked auto-encoders are used for denoising, in our application
this denoising can be interpreted as taking conditional
expectations.
\item The network in \eqref{optimal network parameter} is
fitted with gradient descent, and if the first model $\mu(\bx)$
is also a network with the same architecture, we propose to initialize
gradient descent optimization of \eqref{optimal network parameter}
precisely with the network weights of the first model $\mu(\bx)$.
\item For stochastic gradient descent, one should randomize to order of the indices $1\le l \le 3n$
  in \eqref{optimal network parameter}, so that
  all random mini-batches have  instances of all three kinds.
\item We mask the input data $\bx^{(\bm)}_{l-2n,{\cal C}_l}$, see item
(c) above, i.e., every component of $\bx_l$ is masked independently from the
others with 
probability 1/2. This precisely corresponds to selecting each subset
${\cal C} \subseteq {\cal Q}$ in the SHAP computation \eqref{KernelSHAP} 
with equal probability, which results in $2^q$ subsets of equal 
probability. Alternatively, equivalently,
one could also use a drop-out layer after the input with a
drop-out probability of 1/2. However,
this drop-out approach provides different difficulties.
Drop-out uses the mask value 0, and it cannot easily calibrate the extreme
cases of the null model and the full model. Moreover, drop-out may be more
difficult in implementation if one uses entity embeddings for categorical
feature components (because these act simultaneously on
multiple embedding weights). Our example below will use entity embeddings for categorical
feature components.
\end{itemize}
\end{rems}

There remains the discussion of the mask value $\bm \in \R^q$ in \eqref{masked x}.
We start with the case where $\bx$ only has continuous components. In that case, it may happen
that there is a feature $\bx$ that takes the same value as
the mask $\bm \in \R^q$, i.e., $\bx=\bm$. In the fully masked case we should obtain the null model
\begin{equation*}
  \mu_0 = \mu_{\emptyset}(\bx) ~\stackrel{\rm !}{=} ~ {\rm NN}_{\widehat{\bvartheta}}(\bm),
\end{equation*}
i.e., the estimated network ${\rm NN}_{\widehat{\bvartheta}}$ can perfectly replicate the null model if the feature
$\bX=\bm$ is fully masked. At the same time for the (non-masked) feature value $\bx=\bm$,
we should obtain in the full model
\begin{equation*}
  \mu(\bx) = \mu(\bm) = \mu_{\cal Q}(\bm)~\stackrel{\rm !}{=} ~ {\rm NN}_{\widehat{\bvartheta}}(\bm).
\end{equation*}
These two requirement do not stay in conflict, if we choose the mask value $\bm \in \R^q$ such that
\begin{equation}\label{no conflict}
  \mu(\bm) = \mu_0,
\end{equation}
i.e., we choose the mask value $\bm$ such that the full model has the same prediction in $\bx=\bm$
as the null model (not considering any features).

In practical neural network applications, we typically normalize the continuous feature components
  of $\bx$ to be centered and have
unit variance. This is done to ensure efficient gradient descent fitting. As a consequence, the continuous
feature components of $\bx$ fluctuate around zero. To select the mask value $\bm$ we proceed as follows
in our application below. We choose a small tolerance level $\delta>0$ (we choose $\delta=0.1\%$ in our application),
and we select the mask value $\bm \in \R^q$ as close as possible to the origin among all observed features $(\bx_i)_{i=1}^n$ whose
regression value $\mu(\bx_i)$ differs less than $\delta$ from the null model $\mu_0$. That is,
\begin{equation}\label{choice of mask}
  \bm ~ = ~ \underset{\bx_i:\, |\mu(\bx_i)/\mu_0-1|<\delta}{\arg\min}~ \|\bx_i\|,
\end{equation}
where $\|\cdot\|$ is the Euclidean norm. Having a mask value close to the origin ensures that the mask is in the main
body of the (normalized) distribution of the continuous features. 

\medskip

\begin{rems} \normalfont
\begin{itemize}
\item This proposal \eqref{choice of mask} has some similarity
to the Baseline SHAP presented in Sundararajan--Najmi \cite{Sundar}, where the
masked values are set to a baseline feature value $\bx'$. However, the crucial difference is that we do not explicitly use this baseline feature value in our calculation because we 
perform a full conditional expectation in \eqref{conditional expectation}, but
we only use the mask to indicate the network which variables have
not been observed. In some sense, this is equivalent to ``turning off'' some
input components as in drop-out layers, except that our mask value $\bm$ is 
chosen such that in the fully masked (turned off) case, we rediscover the
null model.
\item The crucial property of the choice of the mask $\bm$ is that it can
both reflect the null model $\mu_0$ and the expected value 
$\mu(\bm)$ in the mask $\bx=\bm$ in the full model, see \eqref{no conflict}.
A mask choice close to zero \eqref{choice of mask} has another
nice interpretation, namely, that values close to zero do not have
any big effects given the affine transformations in neural network layers.
\item Remark that the mask choice \eqref{choice of mask} has provided better models than choosing a (remote) mask value, e.g.,
  $\bm =(2,\ldots, 2)^\top$, in the case of normalized features $\bx$. 
  Theoretically, there should not be any difference
  between this choice and choice \eqref{choice of mask} for large neural networks. However, this latter choice has turned
  out to be more difficult in gradient descent fitting, therefore, we prefer \eqref{choice of mask}. Intuitively, a remote mask value will mean in this
  set-up that the null model is not discovered within the full model, but it
  is rather modeled separately beside the full model.
\end{itemize}
\end{rems}

For the implementation of categorical feature components we use the 
method of entity embedding. Assume that $X_j$
is a categorical variable that takes values in the (nominal) set ${\cal A}_j=\{a_1, \ldots, a_K\}$, i.e., $X_j$ takes
$K$ different levels $(a_k)_{k=1}^K$. For entity embedding, one chooses an embedding dimension $b \in \N$,
and then each level $a_k\in {\cal A}_j$ is assigned an embedding weight $\bbb_k \in \R^b$. That is,
we consider the embedding map
\begin{equation*}
\be: {\cal A}_j \to \R^b, \qquad 
X_j=a_k \mapsto \be(X_j)=\bbb_k;
\end{equation*}
we refer to 
Br\'ebisson et al.~\cite{Brebisson}, Guo--Berkhahn \cite{Guo},
Richman \cite{Richman1, Richman2} and
Delong--Kozak \cite{Delong}. The embedding $\be(X_j)\in \R^b$ is then concatenated with the continuous
components of $\bX$, and this concatenation is used as input to the network. Entity embedding adds
another $Kb$ parameters $[\bbb_1, \ldots, \bbb_K] \in \R^{b\times K}$ to the fitting procedure and these embedding parameters
are also learned during gradient descent network training. In this categorical case, we propose for the masking of $X_j$ to
extend the levels ${\cal A}_j$ by a fictitious level $a_{K+1}$ whose embedding weight is initialized for gradient
descent fitting by $\bbb_{K+1}=0 \in \R^b$.

\section{Example: variable importance}
\label{Example: variable importance}
\subsection{Data, predictive model and conditional expectation network}
We apply this conditional expectation network proposal to the
French motor third party liability (MTPL) claims frequency example studied, e.g., in Charpentier \cite{Charpentier},
Lindholm et al.~\cite{LindholmLindskog} and
W\"uthrich--Merz \cite{WM2023}; this data set is available
through the {\sf R} package {\tt CASdatasets} \cite{CASdatasets}. Listing \ref{learningdata} gives an excerpt of the data.
We use the data pre-processing as described
in Chapter 13 of \cite{WM2023}, and we choose the same network architecture as presented
in Example 7.10 of \cite{WM2023}, with entity embeddings
of dimension $b=2$ for the categorical features
{\tt VehBrand} and {\tt Region}, see Listing
7.4 of \cite{WM2023}. We fit this network architecture to the available
data (using early stopping) which gives us the expected
frequency regression function $\bx \mapsto \mu(\bx)$. Note that this
data describes a claims frequency example, which is usually
modeled with a Poisson rate regression.
Therefore, we
use the Poisson deviance loss, denoted by $L$, for model fitting and evaluation. The time exposures are used as weights in model fitting and evaluation; for the Poisson deviance
loss we refer to Example 2.24 in \cite{WM2023}.

\begin{lstlisting}[float=h,frame=tb,caption={Excerpt of the 
French MTPL claims frequency data set.},
label=learningdata]
'data.frame':   678007 obs. of  12 variables:
 $ IDpol     : num  1 3 5 10 11 13 15 17 18 21 ...
 $ Exposure  : num  0.1 0.77 0.75 0.09 0.84 0.52 0.45 0.27 0.71 0.15 ...
 $ Area      : Factor w/ 6 levels "A","B","C","D",..: 4 4 2 2 2 5 5 3 3 2 ...
 $ VehPower  : int  5 5 6 7 7 6 6 7 7 7 ...
 $ VehAge    : int  0 0 2 0 0 2 2 0 0 0 ...
 $ DrivAge   : int  55 55 52 46 46 38 38 33 33 41 ...
 $ BonusMalus: int  50 50 50 50 50 50 50 68 68 50 ...
 $ VehBrand  : Factor w/ 11 levels "B1","B2","B3",..: 9 9 9 9 9 9 9 9 9 9 ...
 $ VehGas    : Factor w/ 2 levels "Diesel","Regular": 2 2 1 1 1 2 2 1 1 1 ...
 $ Density   : int  1217 1217 54 76 76 3003 3003 137 137 60 ...
 $ Region    : Factor w/ 22 levels "R11","R21","R22",..: 18 18 3 15 15 8 8 20 20 12 ...
 $ ClaimNb   : num  0 0 0 0 0 0 0 0 0 0 ...

\end{lstlisting}

\begin{table}[htb!]
\centering
{\small
\begin{center}
\begin{tabular}{|cl||cc|}
\hline
&& \multicolumn{2}{|c|}{Poisson deviance losses} \\
  &  &in-sample & out-of-sample  \\
  &   &on ${\cal L}$     & on ${\cal T}$ \\
\hline\hline
(0)&  null model $\mu_0$ (empirical mean on ${\cal L}$)&25.213& 25.445 \\
  (1)&  full neural network model $\mu(\bx)$ &23.777& 23.849\\\hline
 (2)& approximation ${\rm NN}_{\widehat{\bvartheta}}(\bm)$ of null model &25.213& 25.446\\
 (3)& approximation ${\rm NN}_{\widehat{\bvartheta}}(\bx)$ full model &23.802& 23.847\\\hline
 
   
\end{tabular}
\end{center}}
\caption{Conditional expectation network approximation; 
Poisson deviance losses $L$ in $10^{-2}$.}
\label{hyperparameter 2}
\end{table}

The fitted model $\mu(\bx)$ and its out-of-sample performance are
shown on line (1) of Table \ref{hyperparameter 2}. These results are directly comparable
to Table 7.4 in \cite{WM2023} because we use the same learning and test data split
for the sets ${\cal L}$ and ${\cal T}$, respectively; see Listing 5.2 and Table 5.2 in
\cite{WM2023}.\footnote{There are small numerical differences
between the results on line (1) of Table \ref{hyperparameter 2} and Table 7.4 in \cite{WM2023}
because here we have standardized the continuous feature components to be centered and have unit
variance, whereas as in \cite{WM2023} we have used the MinMaxScaler for pre-processing
the continuous feature components, see (7.29)-(7.30) of \cite{WM2023}. Often standardization
provides slightly superior results over the MinMaxScaler pre-processing.} Line (0)
of Table \ref{hyperparameter 2} shows the null model which uses the empirical mean
for $\mu_0$. The relative increase in out-of-sample Poisson deviance loss $L$ on ${\cal T}$
when moving from the full to the null model is
\begin{equation}\label{total increase}
  \frac{\frac{1}{n}\sum_{i=1}^n L(Y_i, \mu_0)}{\frac{1}{n}\sum_{i=1}^n L(Y_i, \mu(\bx_i))}-1
  ~=~\frac{25.445}{23.849}-1~=~6.70\%.
  \end{equation}
  This is a comparably small value which expresses that we work in a low signal-to-noise ratio situation, which
  is rather typical in actuarial problems. This relative increase 
  \eqref{total increase} is the benchmark that we are going to 
  attribute to the different feature components.

\begin{figure}[htb!]
\begin{center}
\begin{minipage}[t]{0.6\textwidth}
\begin{center}
\includegraphics[width=\textwidth]{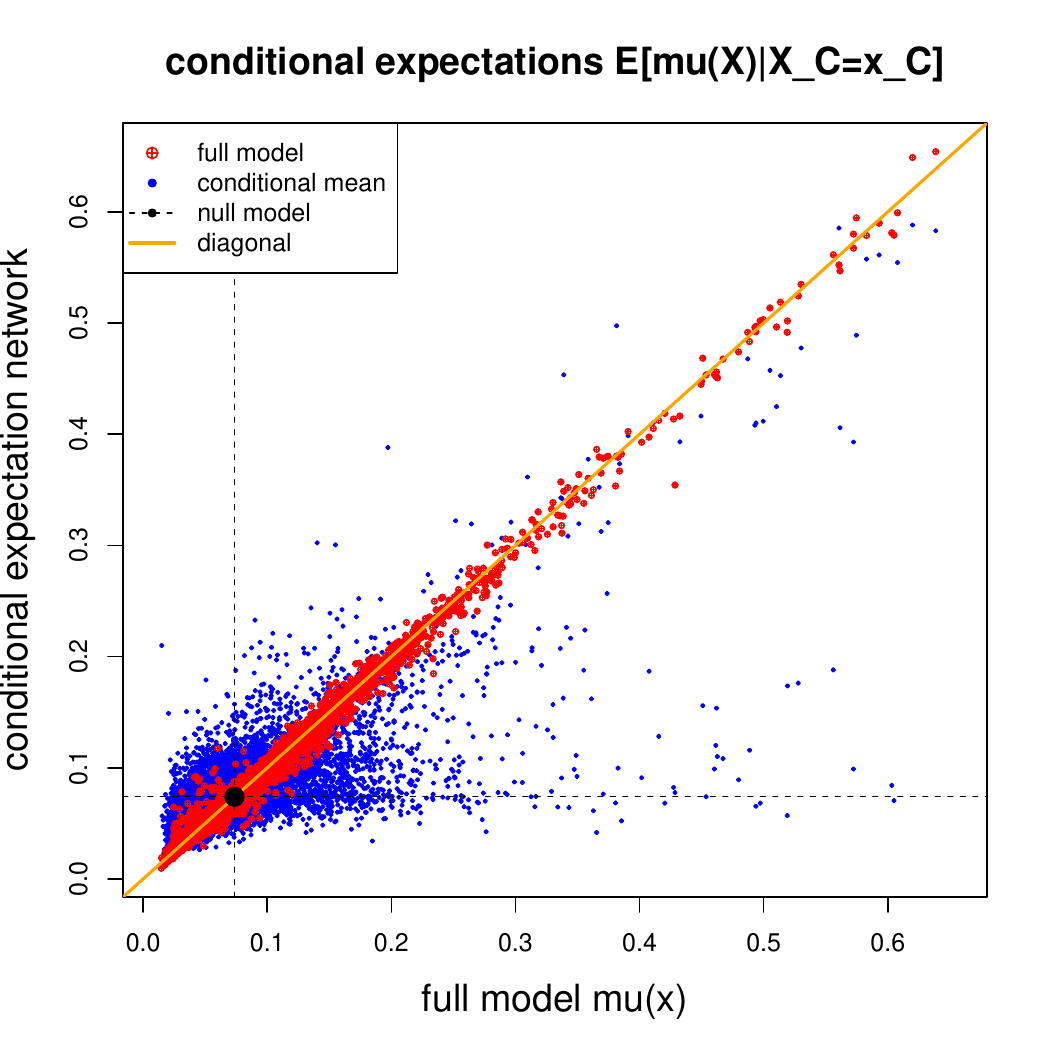}
\end{center}
\end{minipage}
\end{center}
\vspace{-.7cm}
\caption{Conditional expectation network fitting results.}
\label{SHAPnet plot}
\end{figure}

We fit the same network architecture for the calculation of the conditional expectations \eqref{conditional expectation}, and we initialize the
gradient descent
model fitting with the weights of the first network $\mu$.
This is precisely done as in \eqref{optimal network parameter} with the random masking as described
in the previous section.
Figure \ref{SHAPnet plot} shows the results.
The red dots in Figure \ref{SHAPnet plot} reflect the approximation
of the full model $\mu(\bx_i)$, and a perfect network
fit of ${\rm NN}_{\widehat{\bvartheta}}(\bx_i)$, $1\le i \le n$, will set all red dots
precisely on the diagonal orange line. The black dot reflects the 
approximation of the null
model $\mu_0$, and ${\rm NN}_{\widehat{\bvartheta}}(\bm)$ lies perfectly on the diagonal line. The blue
dots reflect the conditional expectations
$\mu_{{\cal C}_i}(\bx_i)$ estimated by
${\rm NN}_{\widehat{\bvartheta}}(\bx^{(\bm)}_{i,{\cal C}_i})$,
where the components
${\cal C}_i^c={\cal Q}\setminus{\cal C}_i$ have been masked. We observe that these blue dots fluctuate quite wildly, of course,
this is expected as we neglect the information in the components $\bX_{{\cal C}_i^c}$.

Lines (2) and (3) of Table \ref{hyperparameter 2} show the performance of this approximation
${\rm NN}_{\widehat{\bvartheta}}$ in the cases of the full model and of the null model. Note that the
fitting procedure has only taken place on the learning data ${\cal L}$, and 
the disjoint sample ${\cal T}$ is only used for the out-of-sample
performance assessment. We observe some in-sample differences, which is a sign that
the first network $\mu$ is in-sample overfitting, because the out-of-sample performance
is equally good between $\mu$ and ${\rm NN}_{\widehat{\bvartheta}}$. 
This is interpret that the conditional expectation network
${\rm NN}_{\widehat{\bvartheta}}$  is a regularized
version of the first neural network $\mu$, and the masked inputs act as drop-out
regularization; indeed, besides for using the conditional expectation network for explanation, the good out-of-sample performance means that this network can also be used for prediction.
Thus, the fitted
network \eqref{neural network} seems to be a good approximation to the (conditional) means in the cases of the
full and the null model.

\subsection{{\tt drop1} and {\tt anova} analyses}
\label{anova and drop1 analyses}
We perform a {\tt drop1} analysis, i.e., we simply set one column of the
design matrix to the mask value, similar to the one used for GLMs; differences are described at the end of this section. The full model $\mu$, given by \eqref{conditional expectation regression example}, dominates
in convex order any model for $j\in {\cal Q}$
\begin{equation*}
\mu_{{\cal Q}\setminus \{j\}}(\bx)=\E\left[\left.\mu(\bX)\right| \bX_{{\cal Q}\setminus \{j\}}=\bx_{{\cal Q}\setminus \{j\}}
  \right]=\E\left[\left.Y\right| \bX_{{\cal Q}\setminus \{j\}}=\bx_{{\cal Q}\setminus \{j\}}
  \right],
\end{equation*}
where we drop the $j$-th component from the information set. On the out-of-sample data ${\cal T}$, we
analyze the relative increase in Poisson deviance loss $L$ using this more crude regression function;
if we drop all components we obtain \eqref{total increase} for the null model.
We define on the out-of-sample data ${\cal T}$ the {\tt drop1} statistics
for $j\in {\cal Q}$
\begin{equation}\label{drop1 formula}
  {\tt drop1}_j ~=~ \frac{\frac{1}{n}\sum_{i=1}^n L(Y_i, \mu_{{\cal Q}\setminus \{j\}}(\bx_i))}{\frac{1}{n}\sum_{i=1}^n L(Y_i, \mu(\bx_i))}-1.
  \end{equation}

\begin{figure}[htb!]
\begin{center}
\begin{minipage}[t]{0.48\textwidth}
\begin{center}
\includegraphics[width=\textwidth]{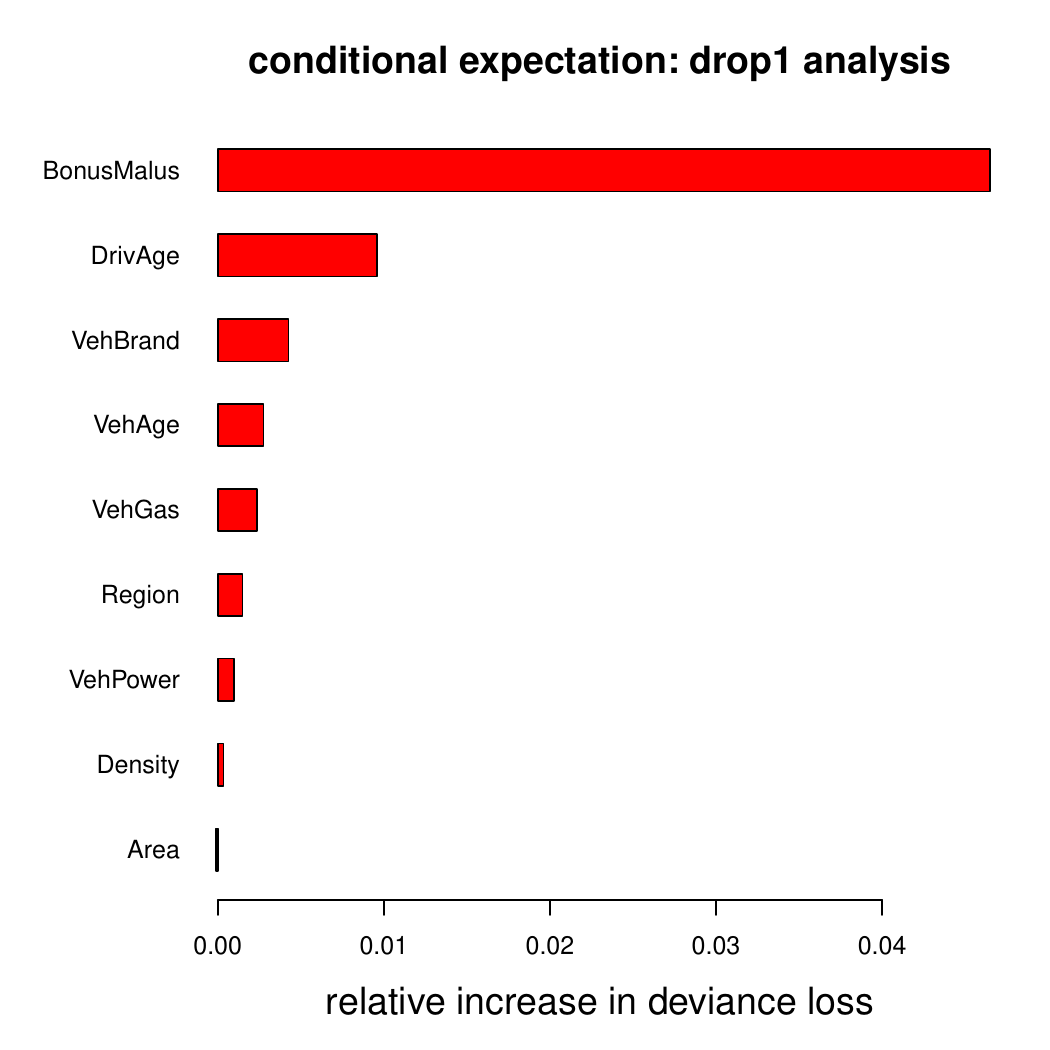}
\end{center}
\end{minipage}
\begin{minipage}[t]{0.48\textwidth}
\begin{center}
\includegraphics[width=\textwidth]{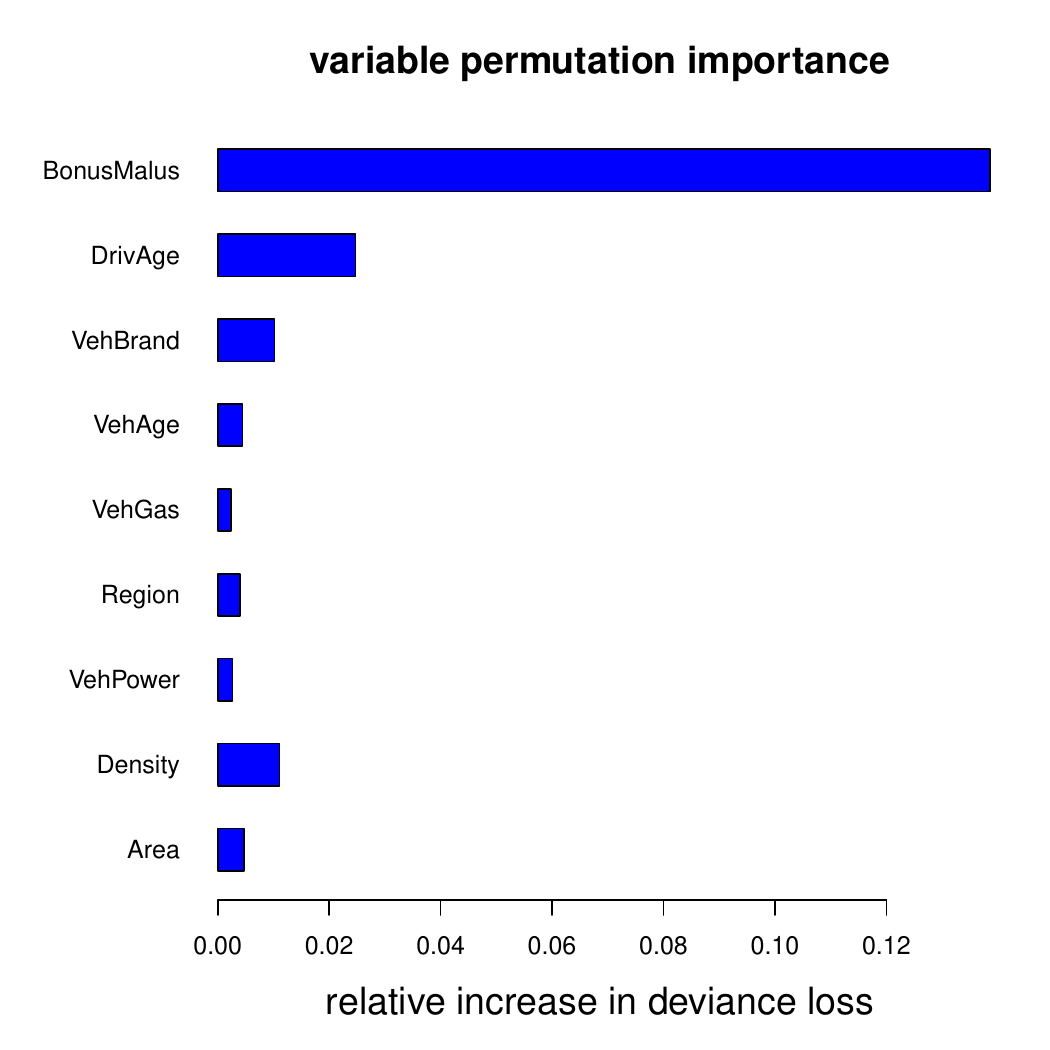}
\end{center}
\end{minipage}
\end{center}
\vspace{-.7cm}
\caption{(lhs) Conditional expectation network for ${\tt drop1}_j$ importances
  of individual components $j \in \{{\tt BonusMalus}, \ldots, {\tt Area}\}$, (rhs)
variable permutation importance (VPI) of Breiman \cite{BreimanRF}; the $x$-scale differs and the order on the $y$-axis is the same.}
\label{VPI plot}
\end{figure}

  Figure \ref{VPI plot} (lhs) shows the results. Dropping the variable {\tt BonusMalus} leads to the biggest
  increase in out-of-sample loss of 4.50\%,
compare to \eqref{total increase},
  and we conclude that this is the most important variable in this prediction problem (using a {\tt drop1} analysis).
  At the other end, {\tt Density} and {\tt Area} do not seem to be important, and may be dropped from the analysis.
  In fact, ${\tt drop1}_{\tt Density}=0.04\%>0$ is slightly positive and ${\tt drop1}_{\tt Area}=-0.01\%<0$ is negative (out-of-sample).
The latter says that we should (clearly) drop the {\tt Area} component from the feature $\bX$, because inclusion of {\tt Area} negatively impacts the out-of-sample performance.

\begin{figure}[htb!]
\begin{center}
\begin{minipage}[t]{0.48\textwidth}
\begin{center}
\includegraphics[width=\textwidth]{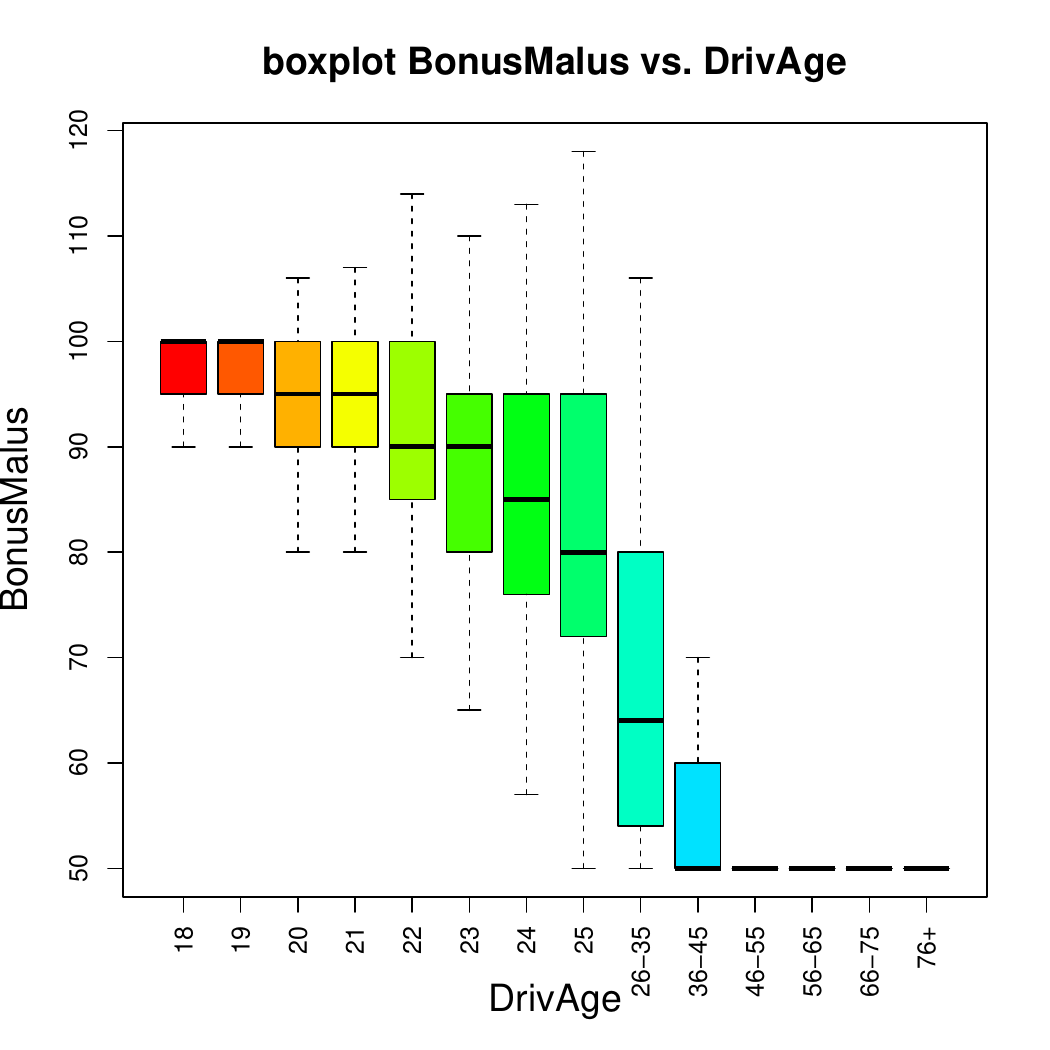}
\end{center}
\end{minipage}
\begin{minipage}[t]{0.48\textwidth}
\begin{center}
\includegraphics[width=\textwidth]{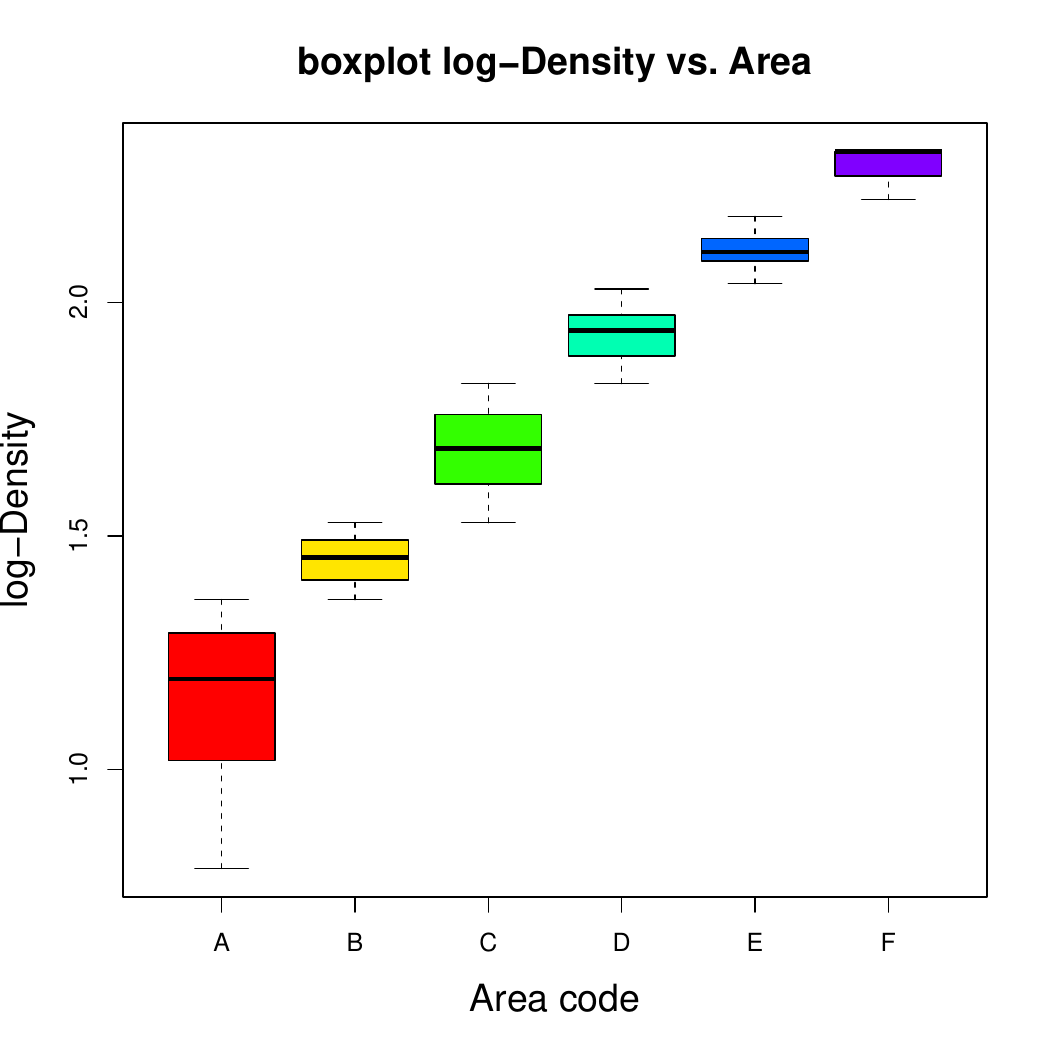}
\end{center}
\end{minipage}
\end{center}
\vspace{-.7cm}
\caption{Dependence between the feature components of $\bX \sim \pi$: (lhs) {\tt DrivAge} and {\tt BonusMalus},
  and (rhs) {\tt Area} code and {\tt Density}; 
  this plot is taken from Figure 13.12 in \cite{WM2023}; the plots only show the whiskers but not the
  outliers.}
\label{colinearity plot}
\end{figure}

We compare these {\tt drop1} importances to the variable permutation importance (VPI) figures of 
Breiman \cite{BreimanRF}. VPI randomly permutes one component of the features $(\bx_i)_{i=1}^n$ at a time across the entire sample, and
then studies the change in out-of-sample loss compared to the full model.
The corresponding results are shown in Figure \ref{VPI plot} (rhs);
we keep the same order on the $y$-axis. We observe bigger magnitudes and also
a slightly different ordering compared to the {\tt drop1} analysis. The difficulty with the VPI analysis
is that it does not properly respect the dependence structure between the feature components of $\bX \sim \pi$, e.g.,
if two feature components are colinear we cannot randomly permute one 
component across the entire portfolio without
changing the other one correspondingly. In Figure \ref{colinearity plot} we show the existing
dependence in our example between {\tt DrivAge} and {\tt BonusMalus} (lhs) and between {\tt Area}
and {\tt Density} (rhs). For instance, we cannot change the {\tt Area} code from A to D without changing
{\tt Density} correspondingly. This is precisely done in our {\tt drop1} analysis using the conditional
expectations $\mu_{{\cal Q}\setminus \{j\}}(\bx)$, but it is not done in VPI. Therefore, the 
conditional expectation results are more reliable to measure variable importance
in this case.

Moreover, since there are no young car drivers (below age 25) who are on the lowest bonus-malus level of 50\%, 
see Figure \ref{colinearity plot} (lhs), the
(fitted) regression function $\bx\mapsto\mu(\bx)$ is not (well-)specified for such feature values, in fact it is undefined on this part of the feature
space. Therefore, we can 
extrapolated $\mu$ arbitrarily to this part of the feature space because this extrapolation cannot be back-tested on data.
Precisely this problem questions the magnitudes in the VPI plot, because
different extrapolations give us
different magnitudes of increases in deviance losses.

\begin{figure}[htb!]
  \begin{center}
\begin{minipage}[t]{0.48\textwidth}
\begin{center}
\includegraphics[width=\textwidth]{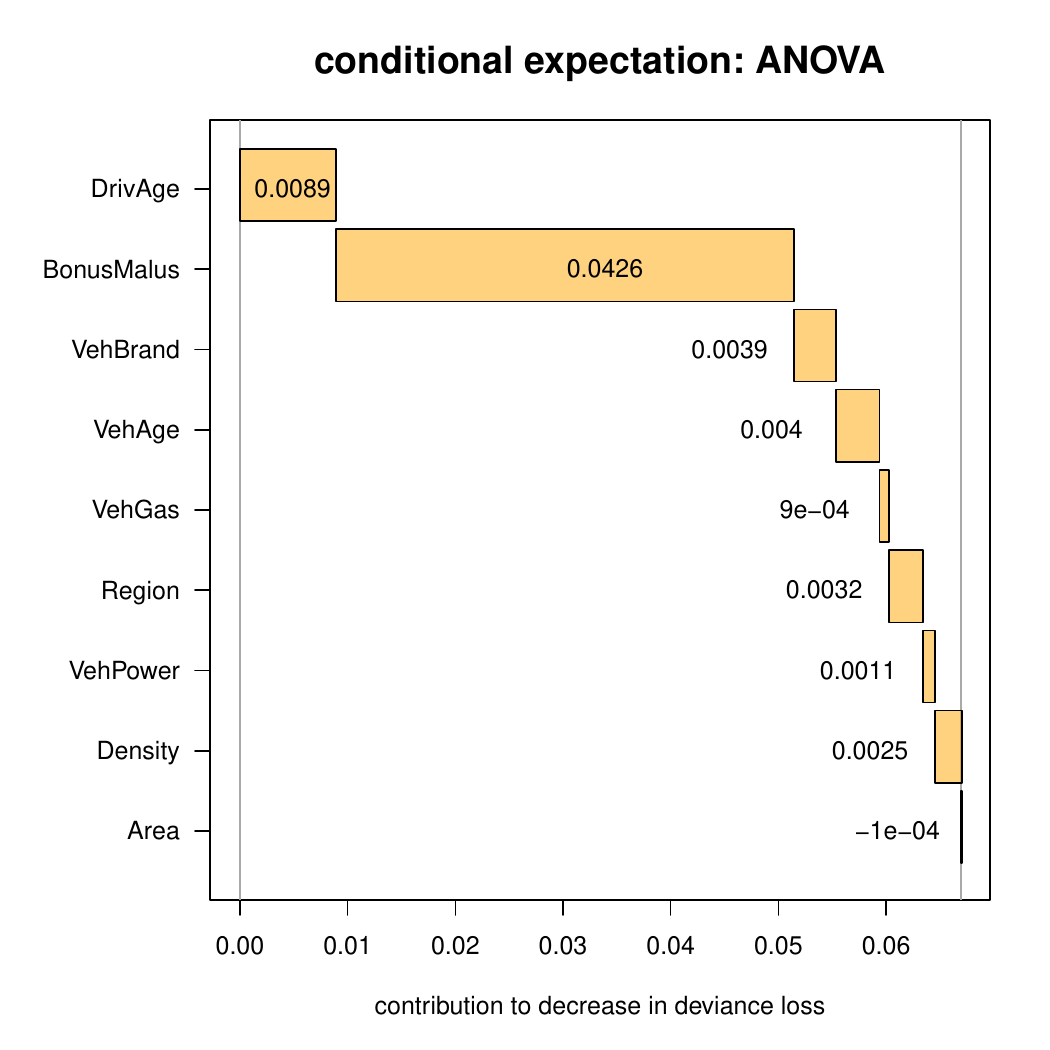}
\end{center}
\end{minipage}
    
\begin{minipage}[t]{0.48\textwidth}
\begin{center}
\includegraphics[width=\textwidth]{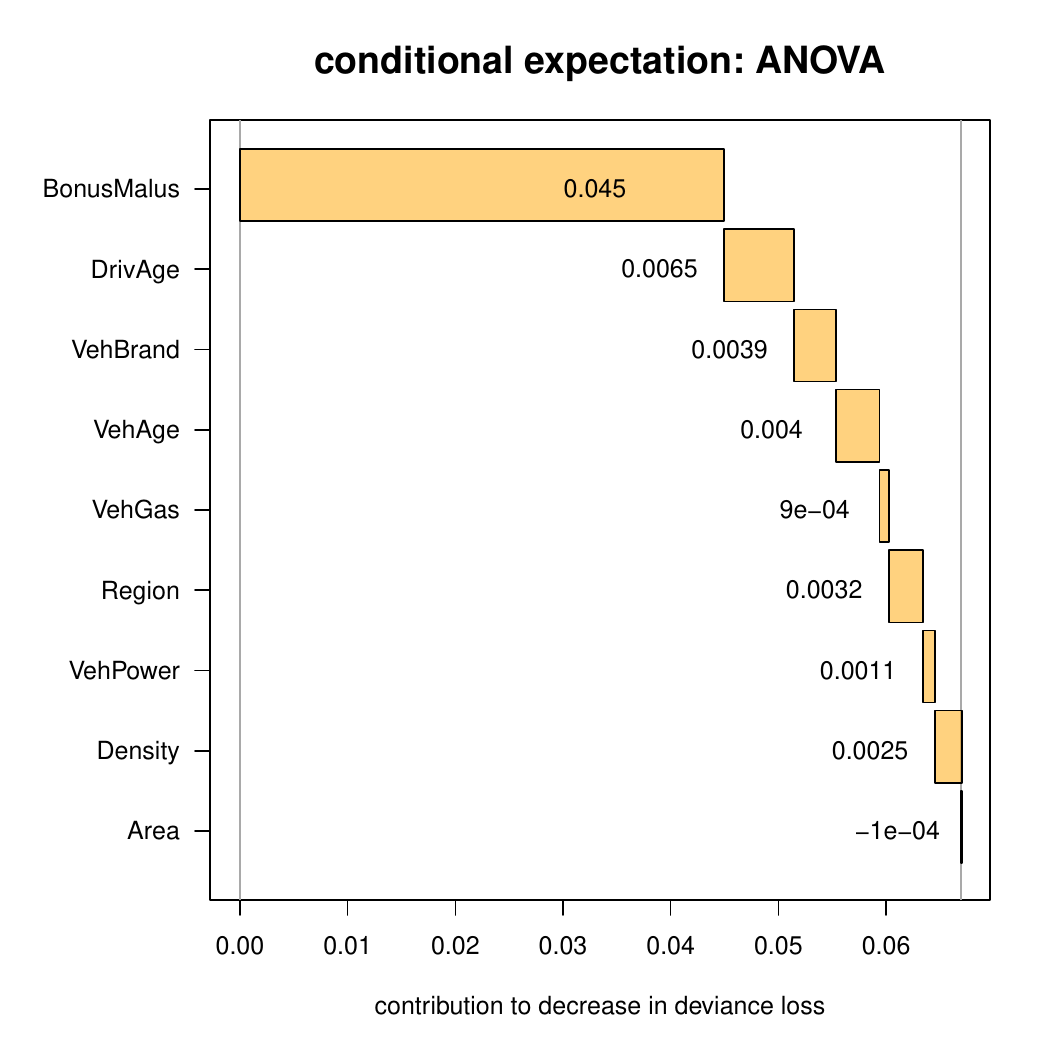}
\end{center}
\end{minipage}
\begin{minipage}[t]{0.48\textwidth}
\begin{center}
\includegraphics[width=\textwidth]{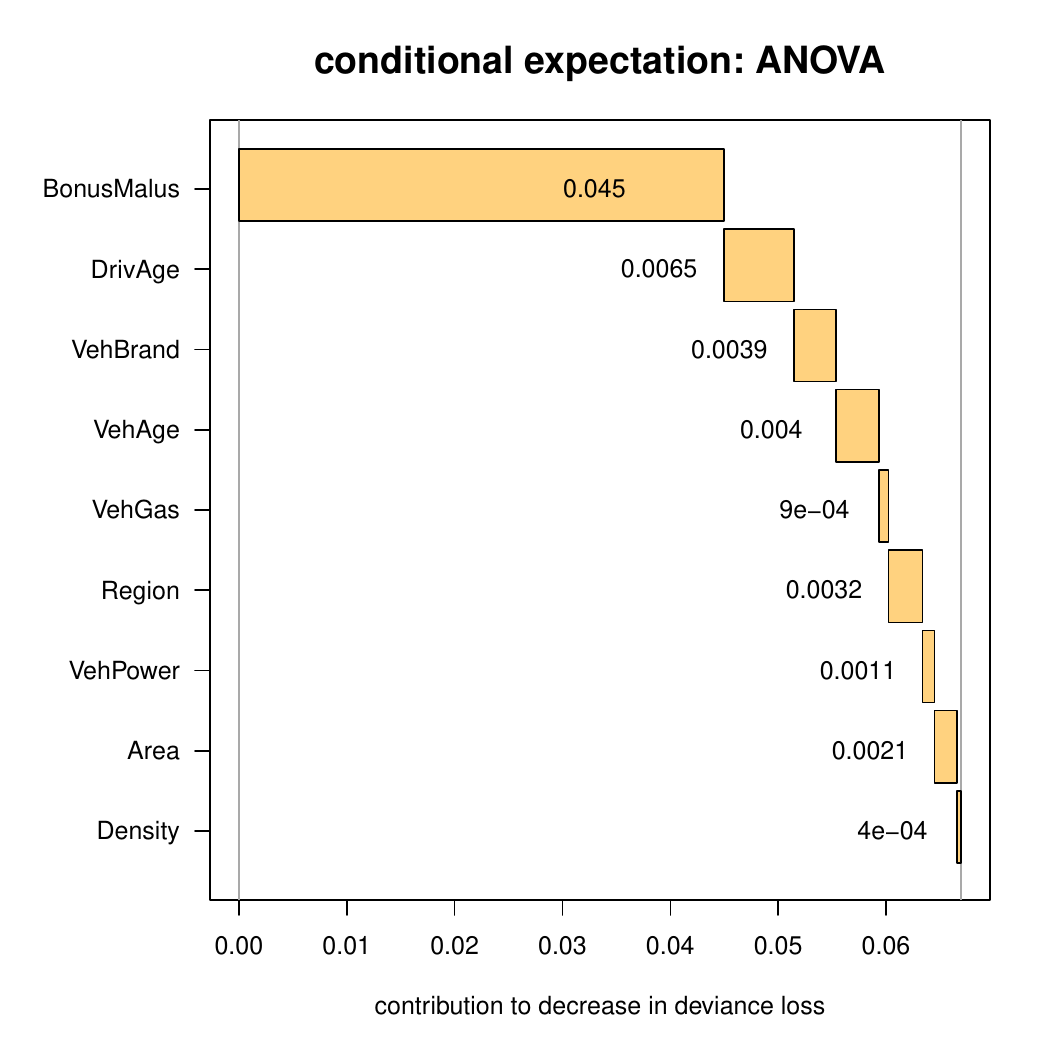}
\end{center}
\end{minipage}
\end{center}
\vspace{-.7cm}
\caption{{\tt anova} analyses for different orderings 
for $j\in {\cal Q}$ of the feature components.}
\label{ANOVA}
\end{figure}

\medskip

Next, we study an {\tt anova} analysis, similarly to the one offered in the {\sf R} package for GLMs;
we also refer to Section 2.3.2 in McCullagh--Nelder \cite{MN}. The {\tt anova}
analysis recursively adds feature components to the regression model, and analyzes the change of (out-of-sample) deviance
loss provided by the inclusion of each new feature component. To have the same units as in the previous
analyses we scale these changes of deviance losses with the loss of the full model, precisely as in \eqref{total increase} and
\eqref{drop1 formula}. The {\tt anova} analysis then gives us an additive decomposition of
the total gap in losses between the full and null model, i.e., it explains the 
difference of 6.70\% given
in \eqref{total increase}. For this we choose a sequence of
mutually different indices $j_1, \ldots, j_q \in {\cal Q}$, and we define the contribution of feature component $X_{j_k}$ to the decrease
in deviance loss by
\begin{equation}\label{anova formula}
  {\tt anova}_{j_k} ~=~ \frac{\frac{1}{n}\sum_{i=1}^n L\left(Y_i, \mu_{\{j_1,\ldots, j_{k-1} \}}(\bx_i)\right)-
    L\left(Y_i, \mu_{\{j_1,\ldots, j_{k} \}}(\bx_i)\right)}{\frac{1}{n}\sum_{i=1}^n L(Y_i, \mu(\bx_i))},
  \end{equation}
  for $k=1,\ldots, q$; for the first term $k=1$ in \eqref{anova formula} 
  we set $\mu_{\emptyset}(\bx)=\mu_0$.
  This gives us an additive decomposition of \eqref{total increase}, i.e.,
  \begin{equation*}
    \sum_{k=1}^q {\tt anova}_{j_k} ~=~
  \frac{\frac{1}{n}\sum_{i=1}^n L(Y_i, \mu_0)}{\frac{1}{n}\sum_{i=1}^n L(Y_i, \mu(\bx_i))}-1
  ~=~\frac{25.445}{23.849}-1~=~6.70\%.
  \end{equation*}  
  Moreover, we have ${\tt anova}_{j_q}={\tt drop1}_{j_q}$, this is the
  last included component $j_q \in {\cal Q}$. Figure \ref{ANOVA} (top) shows the waterfall graph of the {\tt anova}
  analysis stating the corresponding decreases in losses in the plot, e.g., ${\tt anova}_{j_1}=0.89\%$ for $j_1={\tt DrivAge}$. We observe that
  by far the biggest contribution is provided by {\tt BonusMalus}, which gives us the same conclusion about variable
  importance as Figure \ref{VPI plot}. The difficulty with the {\tt anova} analysis is that the contributions depend on the
  order $j_1, \ldots, j_q$ of the inclusion of the feature components. In Figure \ref{ANOVA} (bottom-lhs) we exchange
  the order of {\tt DrivAge} and {\tt BonusMalus}, and then ${\tt anova}_{j_2}=0.65\%<0.89\%$ for $j_2={\tt DrivAge}$, because part
  of the decrease of loss has already been explained by {\tt BonusMalus} (through the dependence in $\bX$ and the interactions in the regression function
  $\mu$). This is also important for dropping variables. In Figure \ref{VPI plot}, the two variables {\tt Density} and {\tt Area}
  provide the smallest values for ${\tt drop1}_j$. However, we also know that these two components are almost colinear,
  see Figure \ref{colinearity plot} (rhs). Figure \ref{ANOVA} (bottom) verifies that one of these two components should be
  included in the model, (bottom-lhs) considers the order {\tt Density}-{\tt Area} and (bottom-rhs) the order {\tt Area}-{\tt Density}.
  Integrating the first of these two variables in position $q-1$
  into the model gives a higher contribution than the two other variables
  {\tt VehPower} or {\tt VehGas}, being considered first. Therefore, {\tt Density}/{\tt Area} is more important than these
  two other variables, but integration of one of them is sufficient.

\medskip

{\bf Discussion and comparison to {\tt drop1} and {\tt anova}
analyses in GLMs.}\\
There is an important difference between the {\tt anova} analysis offered
by GLMs and our {\tt anova} analysis. We discuss this. The {\tt anova}
analysis for GLMs compares (two) nested GLMs. The null hypothesis states that
the smaller GLM is sufficient against the alternative hypothesis that
we should work in a bigger GLM including more components.
Having nested GLMs, this hypothesis can be tested using a likelihood
ratio test (LRT). The starting point of this test is the hypothesis that the smaller GLM
is sufficient.

Our {\tt anova} analysis starts from the bigger model that
specifies a regression function $\bx \mapsto \mu(\bx)$, and the smaller
models are obtained by considering conditional expectations of that
regression function. This provides (a sort of) nested models, but typically
these nested models will not be of the same type. E.g., if we start
with a GLM with log-link function we have regression function
\begin{equation*}
\bx ~\mapsto ~ \mu(\bx)=\exp \left\{ \beta_0 + \sum_{j=1}^q \beta_j x_j \right\},
\end{equation*}
for regression parameter $\bbeta =(\beta_0,\ldots, \beta_q)^\top \in 
\R^{q+1}$. If we drop the last component we receive
\begin{eqnarray*}
\mu_{{\cal Q}\setminus \{q\}}(\bx)&=&
\E \left[ \left. \mu(\bX) \right|\bX_{{\cal Q}\setminus \{q\}}
=\bx_{{\cal Q}\setminus \{q\}} \right]
\\&=&
\exp \left\{ \beta_0 + \sum_{j=1}^{q-1} \beta_j x_j \right\}
\E \left[ \left. \exp \{ \beta_q X_q \}\right|\bX_{{\cal Q}\setminus \{q\}}
=\bx_{{\cal Q}\setminus \{q\}} \right].
\end{eqnarray*}
This last conditional expectation can have any functional form
in $\bx_{{\cal Q}\setminus \{q\}}$, and we do not necessarily have
a GLM for this reduced model.

\subsection{Marginal conditional expectation plot}
\label{sec: MCEP}
A very popular visual explainability tool in machine learning is the
PDP. The PDP has been introduced and studied by
Friedman \cite{friedman2001greedy} and Zhao--Hastie \cite{zhao2019causal}.
PDPs marginalize the regression function $\mu(\bX)$ for all components $X_j$ of $\bX$, $j\in {\cal Q}$,
by considering an unconditional expectation
\begin{equation}\label{PDP formula}
  x_{j}~\mapsto~ \E \left[ \mu\left(\bX_{{\cal Q}\setminus \{j\}},x_{j}\right) \right].
\end{equation}
The unconditional expectation \eqref{PDP formula} sets the $j$-th component of the feature equal to $x_j$, and 
averages over the remaining feature components $\bX_{{\cal Q}\setminus \{j\}}$ without considering the correct dependence
structure between $\bX_{{\cal Q}\setminus \{j\}}$ and $X_{j}=x_{j}$.
That is, equivalently to VPI in Figure  \ref{VPI plot} (rhs), the true dependence structure
is neglected in \eqref{PDP formula}, and it precisely suffers the same deficiency because we may average over
feature combinations that do not occur in the data, e.g., due to colinearity. This issue is generally criticized
in the literature; see, e.g., Apley--Zhu \cite{Apley}. Based on the estimated conditional expectation network
${\rm NN}_{\widehat{\bvartheta}}$, we can easily correct for this deficiency by considering the
  marginal conditional expectations, for $j\in {\cal Q}$,
  \begin{equation}\label{marginal conditional expectations definition}
x_{j}~\mapsto~    
\mu_{ \{j\}}(\bx)=\E\left[\left.\mu(\bX)\right| X_{ j}=x_{j}
  \right]=\E\left[\left.Y\right| X_{ j}=x_{j}
  \right].
\end{equation}

\begin{figure}[htb!]
\begin{center}
\begin{minipage}[t]{0.48\textwidth}
\begin{center}
\includegraphics[width=\textwidth]{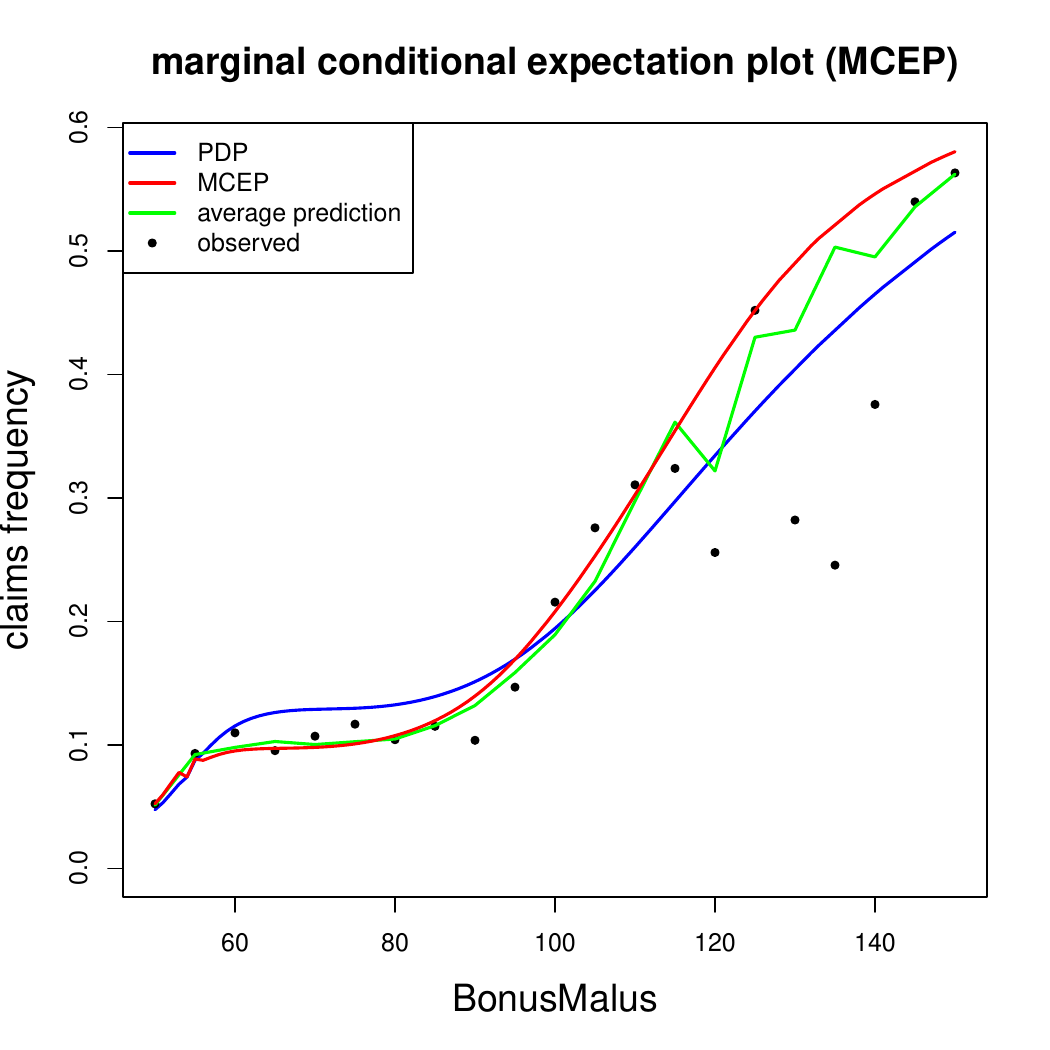}
\end{center}
\end{minipage}
\begin{minipage}[t]{0.48\textwidth}
\begin{center}
\includegraphics[width=\textwidth]{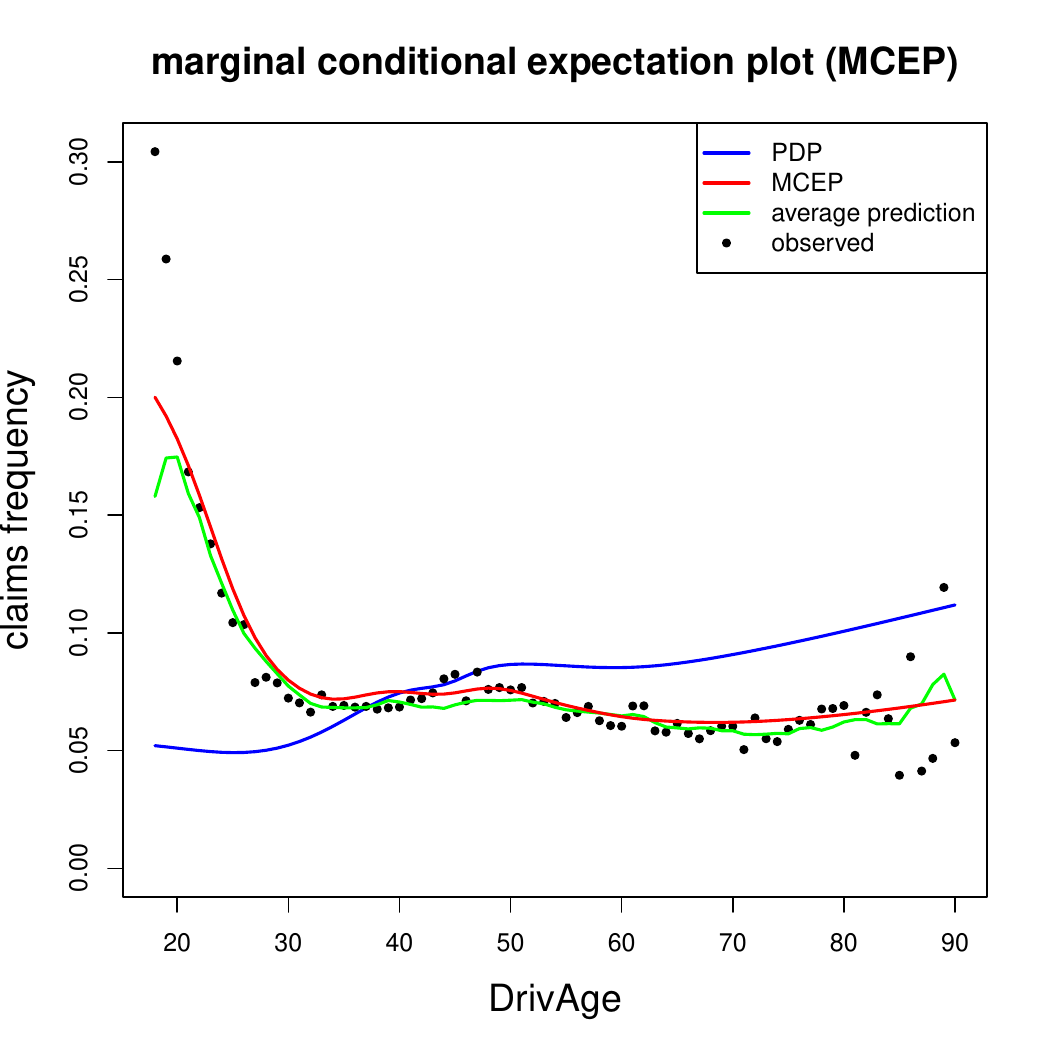}
\end{center}
\end{minipage}
\end{center}
\vspace{-.7cm}
\caption{PDP \eqref{PDP formula} and MCEP \eqref{marginal conditional expectations definition}:
  (lhs) variable {\tt BonusMalus}, and (rhs) variable {\tt DrivAge}.}
\label{plot MCEP}
\end{figure}

Figure \ref{plot MCEP} shows the PDPs \eqref{PDP formula} and the 
MCEPs \eqref{marginal conditional expectations definition} for the two different feature components
$j={\tt BonusMalus}$ on the (lhs) and $j={\tt DrivAge}$ on the (rhs). The blue lines give the PDPs
and the red lines the MCEPs. These lines are complemented by the empirical observations (black dots) and the
average prediction (green lines) given by, respectively,  
\begin{equation*}
\overline{y}_{c_j}=  \frac{\sum_{i=1}^n Y_i\, \mathds{1}_{\{x_{i,j}=c_j\}}}{\sum_{i=1}^n \mathds{1}_{\{x_{i,j}=c_j\}}}
      \qquad \text{ and } \qquad
      \overline{\mu}_{c_j}=\frac{\sum_{i=1}^n \mu(\bx_i)\, \mathds{1}_{\{x_{i,j}=c_j\}}}{\sum_{i=1}^n \mathds{1}_{\{x_{i,j}=c_j\}}},
    \end{equation*}
where $c_j$ runs over all levels that feature component $X_j$ of $\bX$ can take.    
For the most important feature variable, $j={\tt BonusMalus}$, the red and blue lines of the MCEP and PDP look fairly similar.
However, for $j={\tt DrivAge}$, the graphs look quite different for young driver ages. As described in the previous
section, young car drivers cannot be on the lowest bonus-malus level of 50\%, therefore, the
PDP may not give reasonable results for young ages. Contrary, the MCEP can deal with such dependencies
and it is verified by Figure \ref{plot MCEP} (rhs) that the red MCEP curve meets the empirical observations $\overline{y}_{c_j}$ (black dots)
quite well, which says that the marginalized conditional version \eqref{marginal conditional expectations definition}
reflects a one-dimensional regression. The empirical version $\overline{y}_{c_j}$ (black dots) is a noisy version of the average
prediction $\overline{\mu}_{c_j}$ (green line), and also the average prediction and the MCEP curve are rather similar. The
MCEP curve is a conditional expectation \eqref{marginal conditional expectations definition}, whereas the
average prediction $\overline{\mu}_{c_j}$ is a randomized version thereof, considering the empirical (realized) portfolio
distribution of $(\bx_i)_{i=1}^n$.

\section{Example: SHAP}
\label{Example: SHAP}
\subsection{Additive and fair decomposition}
\label{Additive and fair mean decomposition}
We discuss SHAP in this section. SHAP is a popular model-agnostic explainability tool for (complex) regression models, see Lundberg--Lee \cite{lundberg2017} and Aas et al.~\cite{Aas},
but it is also increasingly used to solve other decomposition and attribution problems, e.g., 
a risk allocation example in pension insurance is given by Godin et al.~\cite{Godin}. SHAP is motivated by cooperative game
theory. Shapley \cite{Shapley} stated the following axioms for sharing common gains and costs in an additive and fair way within
a cooperation of $q \ge 2$ players. The {\tt anova} analysis \eqref{anova formula} has provided an additive decomposition of the total loss gap between
the full and the null model, but the {\tt anova} decomposition cannot be considered to be fair, because the order
of the inclusion matters in {\tt anova}, see Figure \ref{ANOVA}.

Assume there exists a value function $\nu$ that maps from the power $\sigma$-algebra of ${\cal Q}$ to the real
line, i.e., 
\begin{equation}\label{value function}
\nu: {\cal C}\subseteq {\cal Q} ~\mapsto~ \nu({\cal C})\in\R.
\end{equation}
This value function $\nu({\cal C})$ measures the contribution of each 
coalition ${\cal C} \subseteq {\cal Q}$ to the total payoff
given by $\nu({\cal Q})$.
Shapley \cite{Shapley} postulated the following four axioms to
be desirable properties of an additive and fair distribution 
$(\mu_j)_{j=1}^q=(\mu^{(\nu)}_j)_{j=1}^q$ of the
total payoff $\nu({\cal Q})$  among the
$q$ players; see also Aas et al.~\cite{Aas}:
\begin{itemize}
	\item[(A1)] {\em Efficiency:} $\nu({\cal Q}) - \nu(\emptyset)=\sum_{j = 1}^q \mu_j$. Set $\mu_0=\nu(\emptyset)$.
	\item[(A2)] {\em Symmetry:} If $\nu({\cal C} \cup \{j\}) = \nu({\cal C} \cup \{k\})$ for every ${\cal C} \subseteq {\cal Q} \setminus\{j, k\}$, then $\mu_j = \mu_k$. 
	\item[(A3)] {\em Dummy player:} If $\nu({\cal C} \cup \{j\}) = \nu({\cal C})$ for every ${\cal C} \subseteq {\cal Q} \setminus\{j\}$, then $\mu_j = 0$.
	\item[(A4)] {\em Linearity:} Consider two cooperative games with value functions $\nu_1$ and $\nu_2$. Then, $\mu_j^{(\nu_1 + \nu_2)} = \mu_j^{(\nu_1)} + \mu_j^{(\nu_2)}$ and $\mu_j^{(\alpha \nu_1)} = \alpha \mu_j^{(\nu_1)}$ for all $1 \le j \le q$ and $\alpha \in \R$.
\end{itemize}
The so-called {\it Shapley values} \cite{Shapley} are the only solution to distribute a total payoff $\nu({\cal Q})$
among the $q$ players so that these four axioms (A1)--(A4) are fulfilled,
and they are given for each $j \in {\cal Q}$ by
\begin{equation}\label{Shapley decomposition}
  \mu_j 
  = \sum_{{\cal C} \subseteq {\cal Q} \setminus\{j\}}\, \frac{|{\cal C}|!\,(q - |{\cal C}| - 1)!}{q!}\, \Big[\nu({\cal C} \cup \{j\}) - v({\cal C})\Big];	
      \end{equation}
      we refer to formula (4) in Lundberg--Lee \cite{lundberg2017}.

      \medskip

There remain two important questions: 
\begin{itemize}
\item[(1)] How should the value function \eqref{value function} be chosen if we translate
the cooperative game theoretic result to regression modeling, meaning
that we would like to ``share'' a prediction $\mu(\bx)$
in a fair and additive way (axioms (A1)-(A4)) among the feature components of $\bx$? 
\item[(2)] How can \eqref{Shapley decomposition}
be calculated efficiently?
\end{itemize}

Item (2) has been answered in Theorem 2 of Lundberg--Lee \cite{lundberg2017}, namely, the Shapley value can be obtain
by solving the following constraint weighted square loss minimization problem
\begin{equation}\label{KernelSHAP}
  \underset{(\mu_j)_{j=1}^q}{\arg\min}~
  \sum_{\emptyset \neq {\cal C} \subsetneq {\cal Q}}\, \frac{q-1}{{q \choose |{\cal C}|}|{\cal C}|(q-|{\cal C}|)}\,
  \left(\nu_0({\cal C}) - \sum_{j \in {\cal C}}\mu_j\right)^2,
  \qquad \text{subject to $\sum_{j=1}^q\mu_{j} =\nu_0({\cal Q})$, }\quad
\end{equation}
where we define $\nu_0({\cal C})=\nu({\cal C}) -\nu(\emptyset)$, and 
where we set $\mu_{0} = \nu(\emptyset)$. This approach is commonly
known as KernelSHAP in the literature, and the term before the 
square bracket in \eqref{KernelSHAP} is called Shapley kernel weight. 
Optimization
\eqref{KernelSHAP} states a convex minimization problem with a linear side constraint
which can be solved with the method of Lagrange. For computing \eqref{KernelSHAP} simultaneously
for different instances (different value functions $\nu$, see also
\eqref{conditional expectation value}, below), a more efficient way is to include the side constraint in a different (approximate) way
by extending the summation in \eqref{KernelSHAP} by the term ${\cal C}={\cal Q}$. This extension
gives a Shapley kernel weight of $+\infty$, and to deal with this
undefined value, one simply sets the Shapley kernel weight for the term
${\cal C}={\cal Q}$ to a very large value; see, e.g., Section 2.3.1 of Aas et al.~\cite{Aas}. The optimal solution is
in that case is given by
\begin{equation}\label{efficient Shapley}
  (\mu_j)_{j=1}^q=  \left(Z^\top W  Z\right)^{-1} Z^\top W \bnu,
\end{equation}
with diagonal Shapley kernel weight matrix $W \in \R^{(2^q-1)\times (2^q-1)}$, 
vector $\bnu \in \R^{2^q-1}$ containing all terms $\nu_0({\cal C})
=\nu({\cal C})-\nu({\cal \emptyset})$ of all coalitions $\emptyset \neq {\cal C} \subseteq {\cal Q}$, and design
matrix $Z\in \{0,1\}^{(2^q-1) \times q}$. Note that if one considers different instances (different value functions $\nu$), only the last
term $\bnu$ in \eqref{efficient Shapley} changes, and the remaining terms only need to be calculated once.

The summation in \eqref{KernelSHAP} involves $2^{q}-2$ terms 
which can be large for high-dimensional features $\bX$. Therefore,
in applications, one often uses a randomized version of \eqref{KernelSHAP}
that randomly samples the terms of the summation in \eqref{KernelSHAP} 
with categorical probabilities determined by the Shapley kernel weights; we refer to Section 2.3.1 in Aas et al.~\cite{Aas}.
This solves item (2) from above.

\medskip

Item (1) is more controversial. The Shapley values \eqref{Shapley decomposition} are unique for a given value function
choice \eqref{value function}. Lundberg--Lee \cite{lundberg2017} have proposed to choose as value function the
conditional expectations for a given instance $\bx \in {\cal X}$. That is, for 
a selected $\bx$, we
define the value function
\begin{equation} \label{conditional expectation value}
{\cal C}\subseteq {\cal Q}~\mapsto~
\nu({\cal C}) :=   
\mu_{\cal C}(\bx)=
\E \left[ \left.\mu(\bX) \right| \bX_{\cal C}=\bx_{\cal C} \right],
\end{equation}
see \eqref{conditional expectation}. In the case of tree based regressions, a version of these Shapley values can efficiently
be calculated using the so-called TreeSHAP method of Lundberg et al.~\cite{lundberg2020}.
However, in the general case, there has not been any efficient way of calculating
the conditional expectations \eqref{conditional expectation value} and the Shapley values, respectively.
Therefore, the conditional expectations \eqref{conditional expectation value} have been replaced
by approximations, see formula (11) in Lundberg--Lee \cite{lundberg2017},
\begin{equation} \label{conditional expectation value independence}
\nu({\cal C}) :=   
\E \left[ \mu\left(\bX_{{\cal Q}\setminus {\cal C}},\bx_{\cal C}\right) \right],
\end{equation}
i.e., similarly to VPI in Figure  \ref{VPI plot} (rhs)
and the PDP \eqref{PDP formula}, the true dependence structure
between $\bX_{{\cal Q}\setminus {\cal C}}$ and $\bX_{\cal C}$ is neglected in \eqref{conditional expectation value independence}; sometimes
this is also called interventional SHAP, see Laberge--Pequignot \cite{Laberge}.
In fact, \eqref{conditional expectation value} and \eqref{conditional expectation value independence}
are equal if $\bX_{{\cal Q}\setminus {\cal C}}$ and $\bX_{\cal C}$ are independent.
In our example, this is clearly not the case, see Figure \ref{colinearity plot}. This is also the main
issue raised in Aas et al.~\cite{Aas}, and as an improvement, these authors propose Gaussian approximations to the true dependence structure.
In our example, we directly approximate the conditional expectations using the estimated
network ${\rm NN}_{\widehat{\bvartheta}}$, see \eqref{optimal network parameter}, i.e., we perform a conditional SHAP using the 
surrogate network ${\rm NN}_{\widehat{\bvartheta}}$ for fast
computation.

The concept of using the conditional expectations \eqref{conditional expectation} has bee criticized as a whole in the paper of
Sundararajan--Najmi \cite{Sundar} showing that in some situations this choice leads to unreasonable Shapley
values $(\mu_j)_{j=0}^q$, and these authors propose to use an unconditional
expectation \eqref{conditional expectation value independence} in general. This proposal is also supported
by causal arguments given in Janzing et al.~\cite{Janzing}. However, causal arguments often use strong assumptions
that cannot easily be verified, e.g., the exclusion of unmeasured confounders, and the general use of an
unconditional expectation \eqref{conditional expectation value independence} cannot be supported in situations
like the ones in Figure \ref{colinearity plot}. Namely, in this example, there are no car drivers with
{\tt DrivAge} below 25 having a {\tt BonusMalus} level of 50\%. Therefore, the regression function $\mu(\bx)$
is undetermined for such features $\bx$ and, henceforth, \eqref{conditional expectation value independence}
cannot generally be calculated because the specific value of one variable
leads to constraints in the support of the other variable. This problem can be circumvented by extending the regression function $\mu$
to this part of the feature space, however, this extension is completely subjective because it cannot be
supported  and verified by data. In the examples in the next section, we compare the conditional and unconditional
versions \eqref{conditional expectation value} and \eqref{conditional expectation value independence},
respectively, and for the extrapolation, we simply use the one provided by the fitted neural network.

We remark that there is interesting work that extends Shapley values
to higher order decompositions and representations; we refer to
Tsai et al.~\cite{Tsai} and Hiabu et al.~\cite{Hiabu}. The basic idea
is to give a functional decomposition of the regression function by
including higher interaction terms. This can partly mitigate the difficulty
of the decision whether one should work with conditional or unconditional
expectations, however, some issues remain, e.g., the above mentioned
support constraints cannot be dealt with the (unconstrained) marginal
identification given by formula (2) in Hiabu et al.~\cite{Hiabu}.

\subsection{SHAP for mean decompositions}
\label{SHAP for mean decompositions}
We apply the SHAP explanation to the regression value $\mu(\bx)$ of a given instance $\bx$.
We compare the conditional and unconditional versions 
\eqref{conditional expectation value} and \eqref{conditional expectation value independence}, respectively.
For the unconditional version and its graphical illustrations we use the {\sf R} packages
{\tt kernelshap} \cite{kernelshap} and {\tt shapviz} \cite{shapviz}; we
refer to Mayer et al.~\cite{Mayer2} for more description.

\begin{figure}[htb!]
  \begin{center}
\begin{minipage}[t]{0.49\textwidth}
\begin{center}
\includegraphics[width=\textwidth]{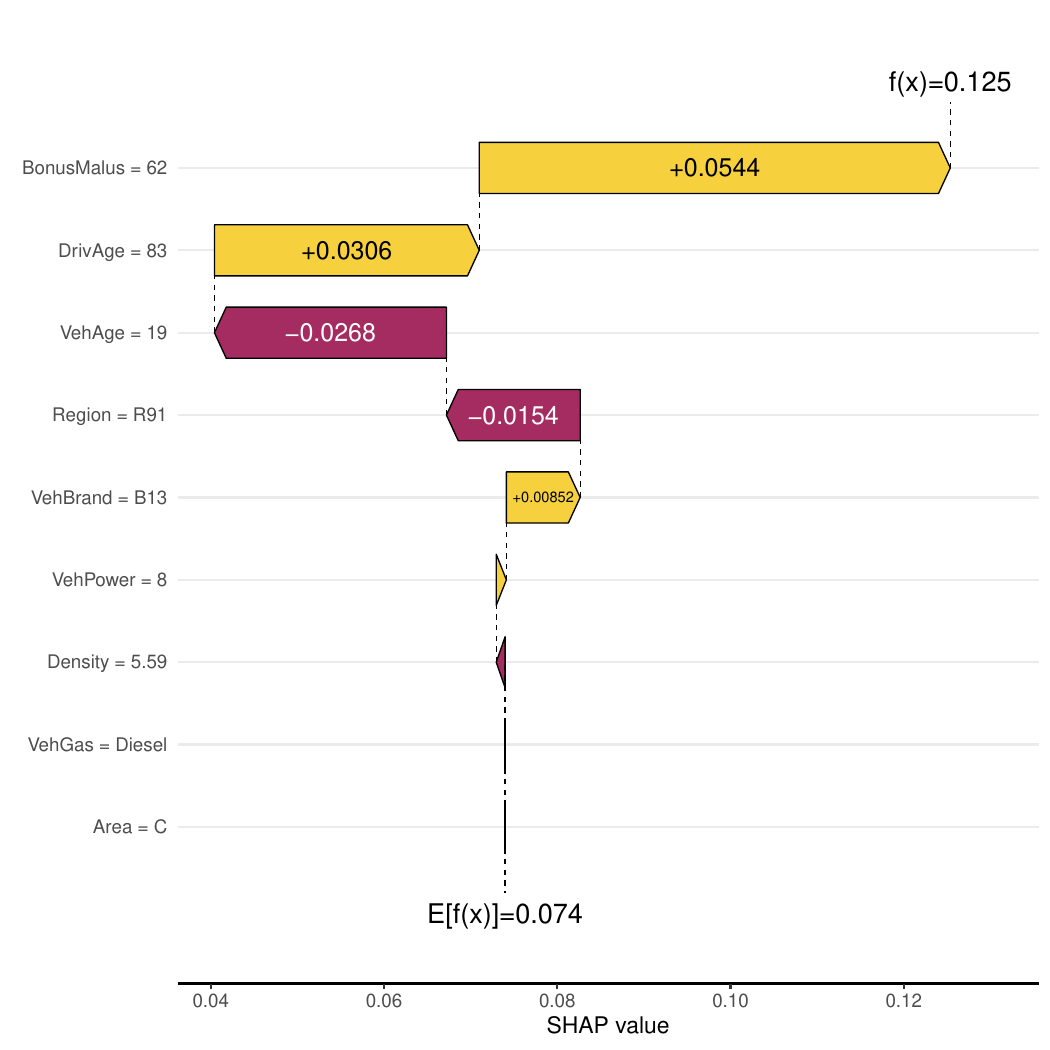}
\end{center}
\end{minipage}
\begin{minipage}[t]{0.49\textwidth}
\begin{center}
\includegraphics[width=\textwidth]{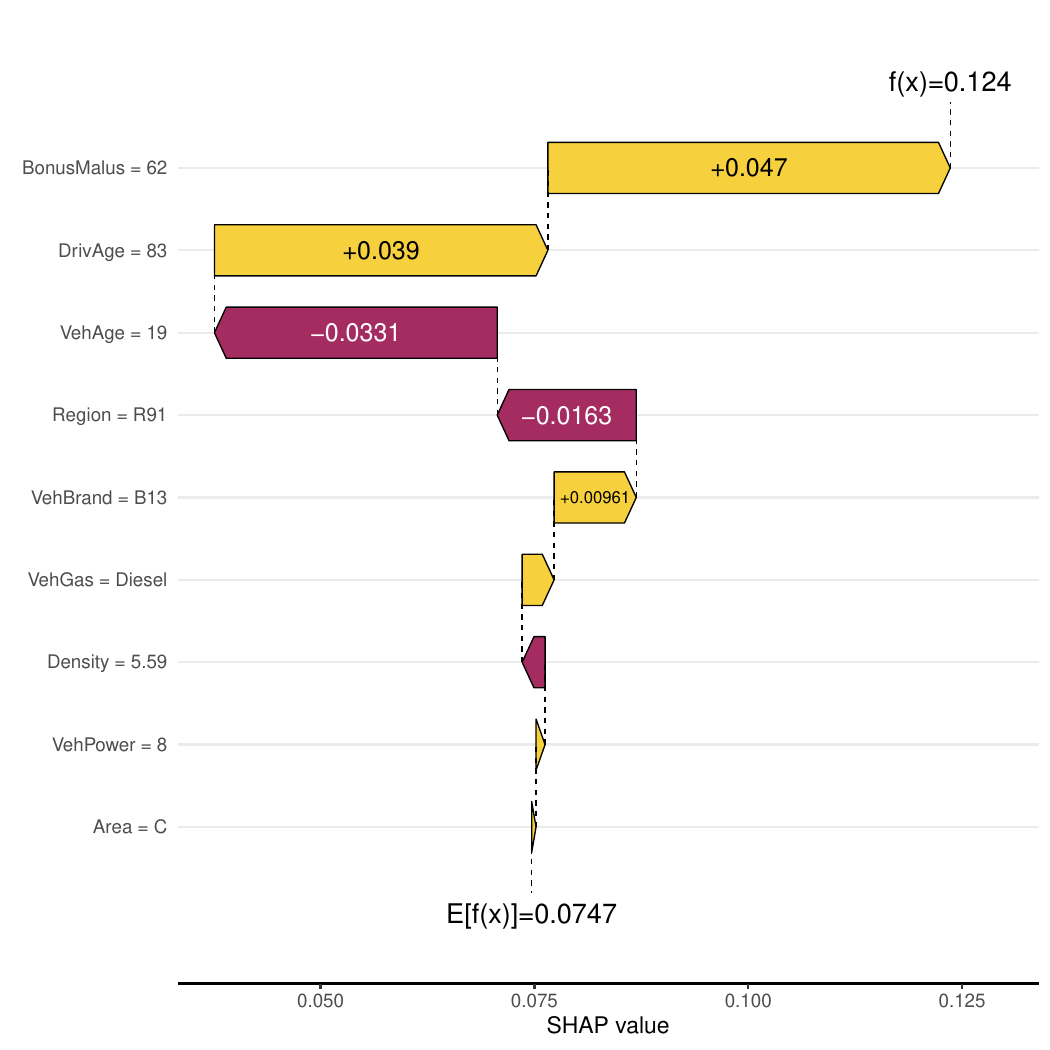}
\end{center}
\end{minipage}
\end{center}
\vspace{-.7cm}
\caption{Waterfall graphs of the Shapley decomposition $(\mu_j)_{j=0}^q$ of $\mu(\bx)$ of a selected instance $\bx \in {\cal X}$: (lhs) conditional expectation
  \eqref{conditional expectation value} for value function $\nu$; (rhs)
unconditional expectation \eqref{conditional expectation value independence} for value function $\nu$;
  these waterfall graphs use {\tt shapviz} \cite{shapviz}.}
\label{Waterfall 1}
\end{figure}

Figure \ref{Waterfall 1} shows the waterfall graphs of the Shapley decomposition $(\mu_j)_{j=0}^q$ of $\mu(\bx)$ of
a given instance with features $\bx \in {\cal X}$; the ordering on the $y$-axis is
according to the sizes of these Shapley values $(\mu_j)_{j=1}^q$. The left-hand side shows the
conditional version \eqref{conditional expectation value} and the right-hand side the unconditional
one \eqref{conditional expectation value independence}. These conditional
SHAP values with \eqref{conditional expectation value} are obtained by using the conditional expectation network
${\rm NN}_{\widehat{\bvartheta}}$ for fast computation. That is, we only
need to fit one single neural network that serves at simultaneously calculating
the conditional expectations of all possible subsets ${\cal C} \subseteq
{\cal Q}$. A naive way would be to fit a network to each subset, which would
require to fit $2^q$ networks.

The results in Figure \ref{Waterfall 1} are rather
similar in this example, and there does not seem to be an issue with the colinearities illustrated in Figure \ref{colinearity plot},
because {\tt Density}/{\tt Area} only has a marginal influence
on  regression value $\mu(\bx)$ and {\tt DrivAge}/{\tt BonusMalus}
is not in the critical (undetermined) part of the feature space.

\begin{figure}[htb!]
  \begin{center}
\begin{minipage}[t]{0.49\textwidth}
\begin{center}
\includegraphics[width=\textwidth]{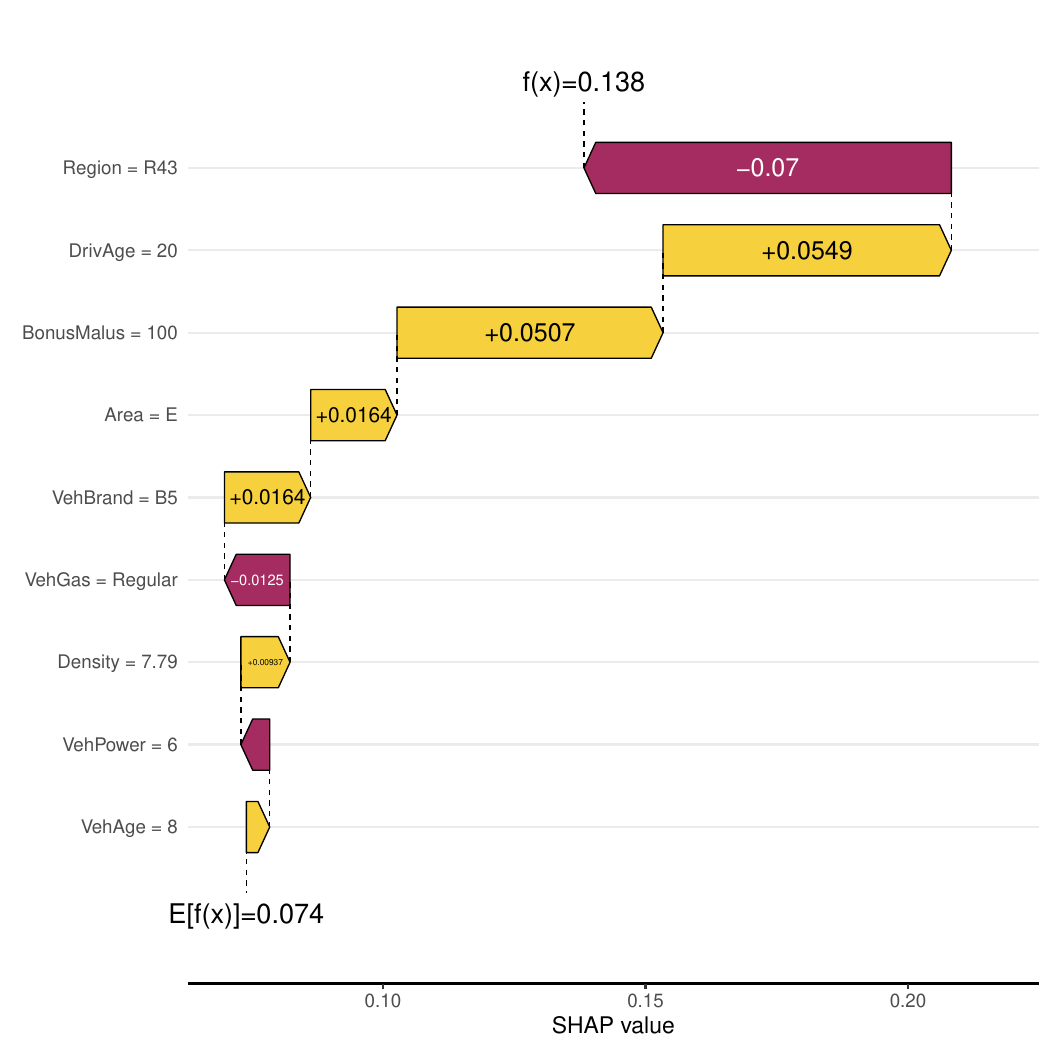}
\end{center}
\end{minipage}
\begin{minipage}[t]{0.49\textwidth}
\begin{center}
\includegraphics[width=\textwidth]{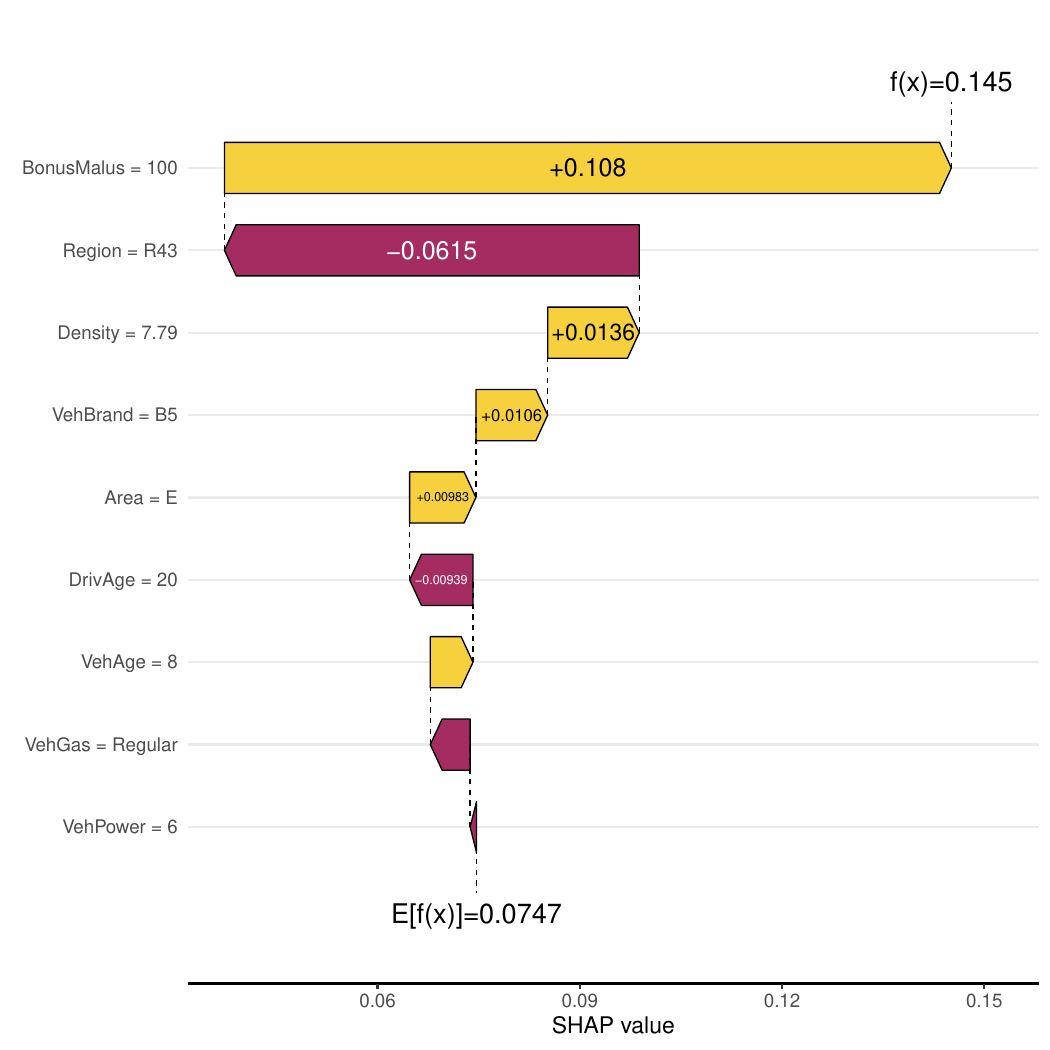}
\end{center}
\end{minipage}
\end{center}
\vspace{-.7cm}
\caption{Waterfall graphs of the Shapley decomposition $(\mu_j)_{j=0}^q$ of $\mu(\bx)$ of a selected instance $\bx \in {\cal X}$: (lhs) conditional expectation
  \eqref{conditional expectation value} for value function $\nu$; (rhs)
  unconditional expectation \eqref{conditional expectation value independence} for value function $\nu$;
  these waterfall graphs use {\tt shapviz} \cite{shapviz}.}
\label{Waterfall 2}
\end{figure}

In Figure \ref{Waterfall 2} we give a second example of a young car driver of age ${\tt DrivAge}=20$. Car drivers enter
a bonus-malus scheme at level 100\%, and every year of accident-free driving decreases this level by 5\% (and an accident
increases the bonus-malus level by a fixed percentage). Thus, it takes at least
10 years of accident-free driving until a car driver can reach the lowest bonus-malus level of 50\%.\footnote{The fact that
  Figure \ref{colinearity plot} does not precisely reflect a 5\% decrease for every accident-free year is an issue in data
  quality, e.g., drivers at the age of 18 can technically not be on a bonus-malus level of 90\%, but there are still a few such
  observations in our data, which we attribute to data error.} As a result, the regression function $\mu$ is undetermined for features having
${\tt DrivAge}=20$ and ${\tt BonusMalus}<90\%$, and we can assign any value to $\mu$ for this feature
as it does not occur in the data. This is precisely what is happening when using the unconditional
version \eqref{conditional expectation value independence} for SHAP, and 
in Figure \ref{Waterfall 2} (rhs) we observe that {\tt BonusMalus}
gets a large attribution if we just extrapolate the (first) neural network
regression function $\mu$ to that part of the feature space. 
Of course, this cannot be justified and supported
by data, as it extrapolates
$\mu$ arbitrarily
to the undefined part of the feature space ${\cal X}$.
In such examples, we give clear preference to the conditional version
 \eqref{conditional expectation value}
on the left-hand side of Figure \ref{Waterfall 2}.

\subsection{LightGBM surrogate model}
We compare the SHAP mean decomposition results of
the previous Section \ref{SHAP for mean decompositions} to the corresponding TreeSHAP results by approximating the full model $\mu(\bx)$ by
a LightGBM surrogate tree regression model.\footnote{To
fit the LightGBM surrogate regression model, we use the same parametrization
as in the model on \url{https://github.com/JSchelldorfer/ActuarialDataScience/tree/master/14 - SHAP}; we also refer to Mayer et al.~\cite{Mayer2}.}
Using this LightGBM surrogate model, we study the resulting TreeSHAP mean
decomposition of Lundberg et al.~\cite{lundberg2020} implemented in
the {\sf R} package {\tt shapviz} \cite{shapviz}.

\begin{figure}[htb!]
  \begin{center}
\begin{minipage}[t]{0.32\textwidth}
\begin{center}
\includegraphics[width=\textwidth]{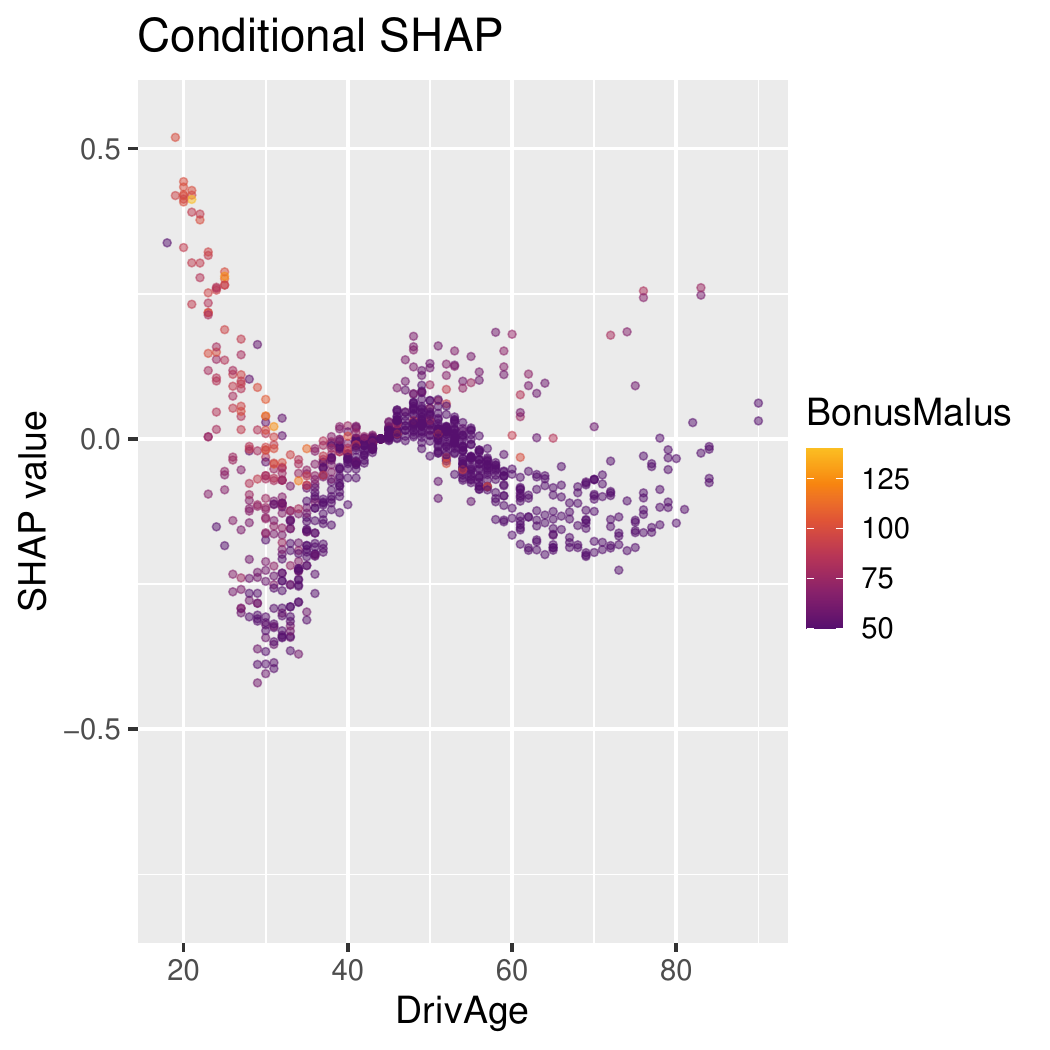}
\end{center}
\end{minipage}
\begin{minipage}[t]{0.32\textwidth}
\begin{center}
\includegraphics[width=\textwidth]{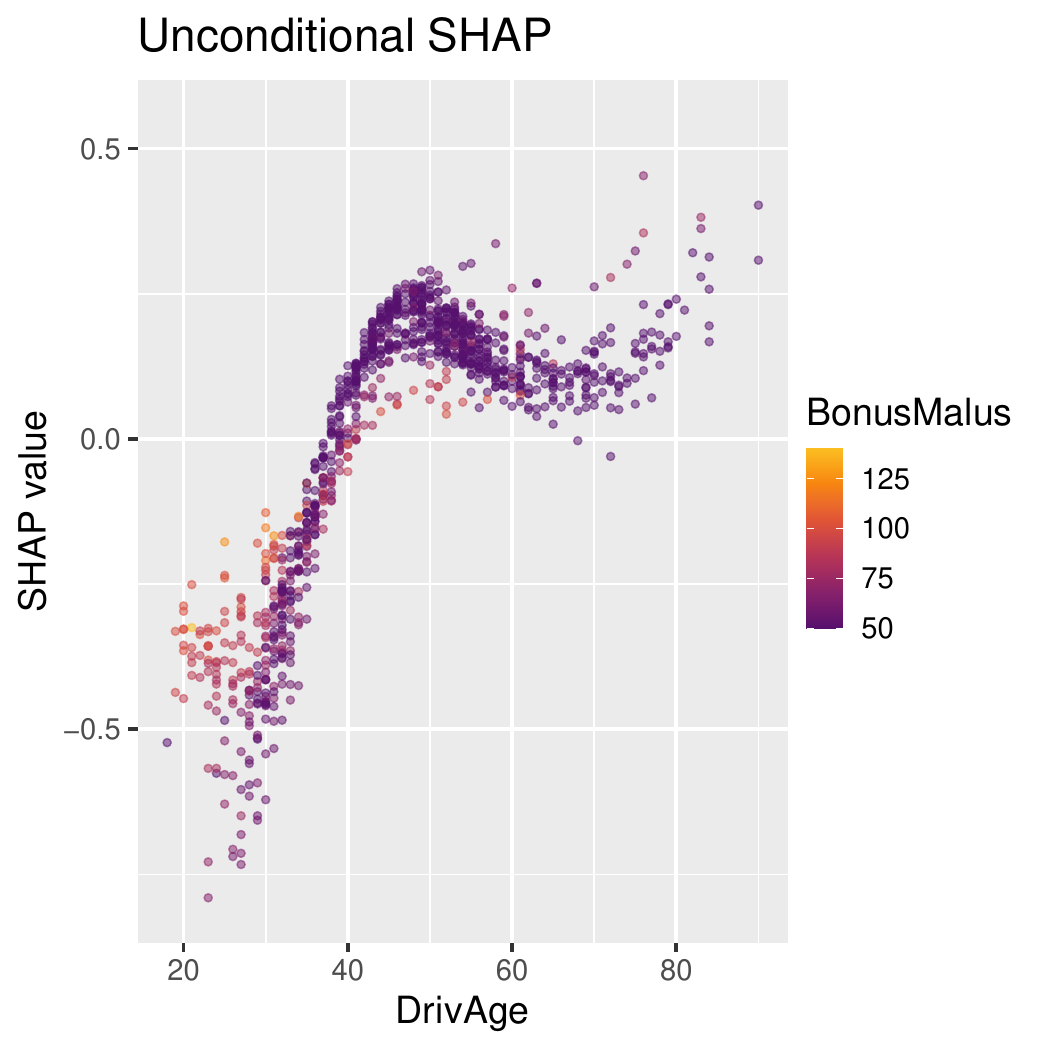}
\end{center}
\end{minipage}
\begin{minipage}[t]{0.32\textwidth}
\begin{center}
\includegraphics[width=\textwidth]{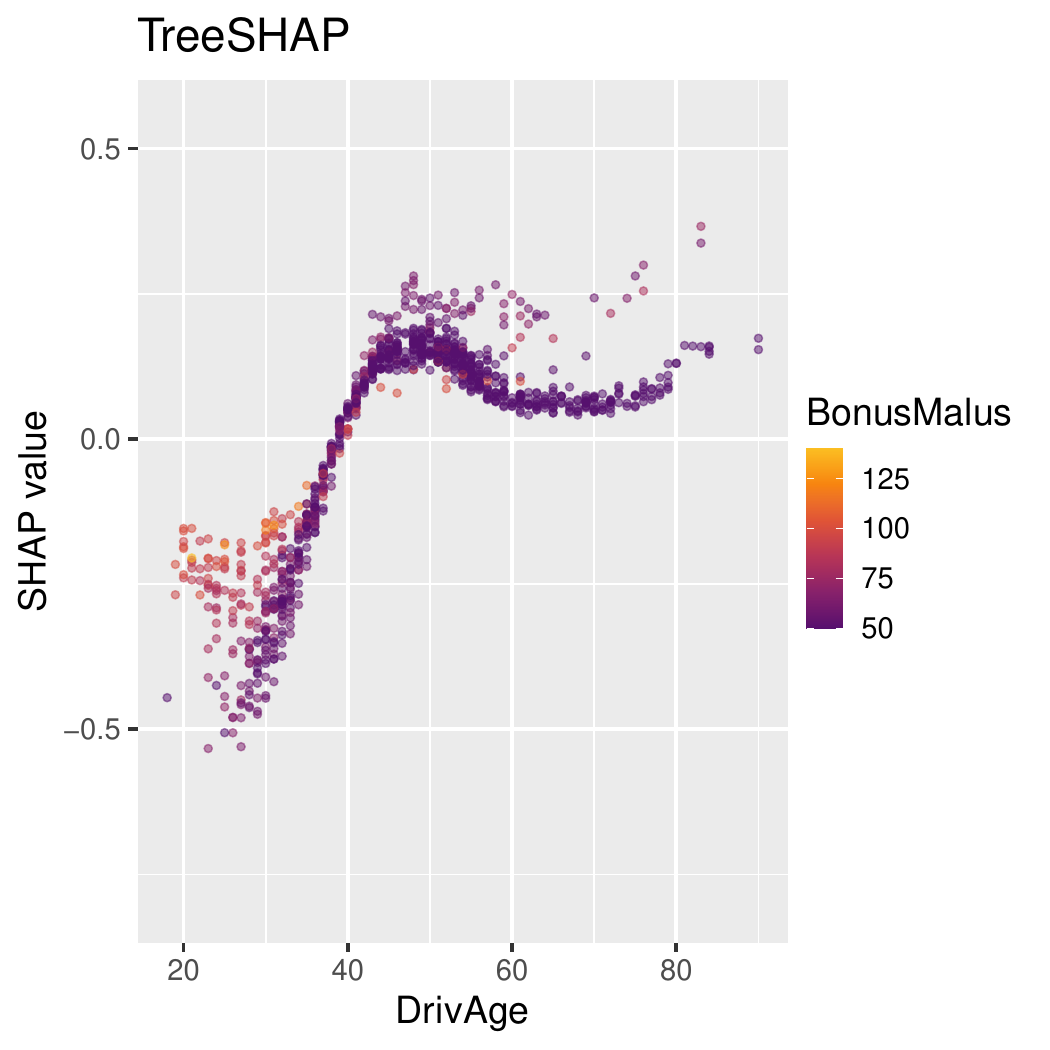}
\end{center}
\end{minipage}

\begin{minipage}[t]{0.32\textwidth}
\begin{center}
\includegraphics[width=\textwidth]{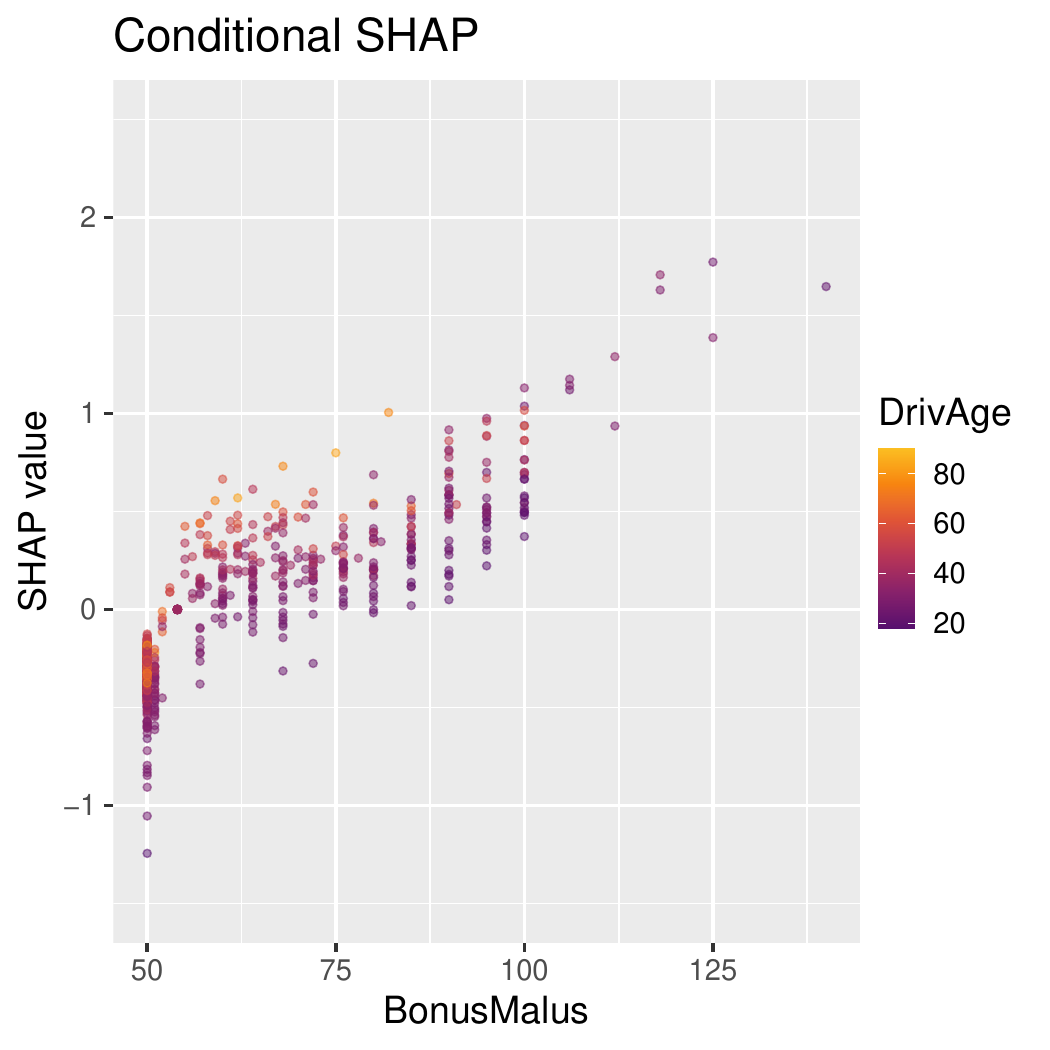}
\end{center}
\end{minipage}
\begin{minipage}[t]{0.32\textwidth}
\begin{center}
\includegraphics[width=\textwidth]{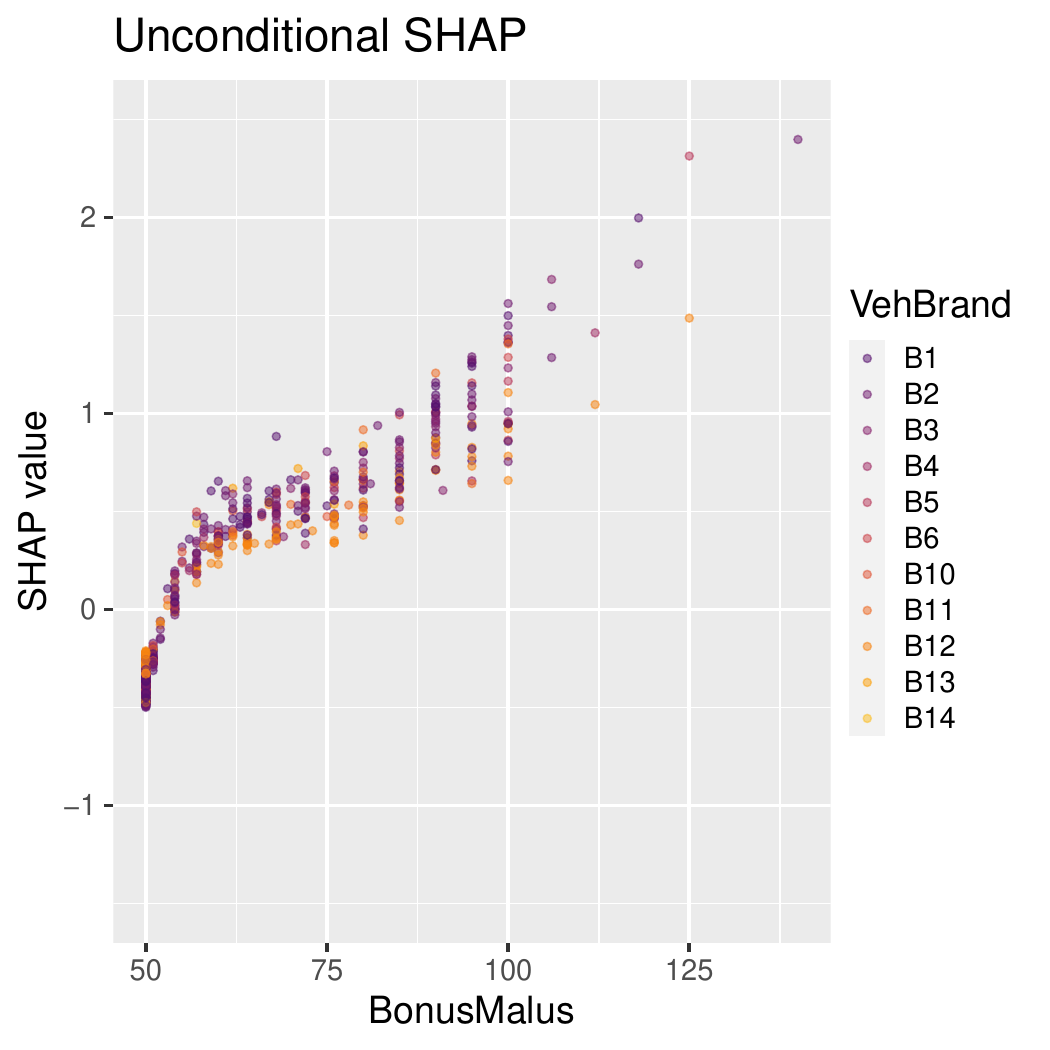}
\end{center}
\end{minipage}
\begin{minipage}[t]{0.32\textwidth}
\begin{center}
\includegraphics[width=\textwidth]{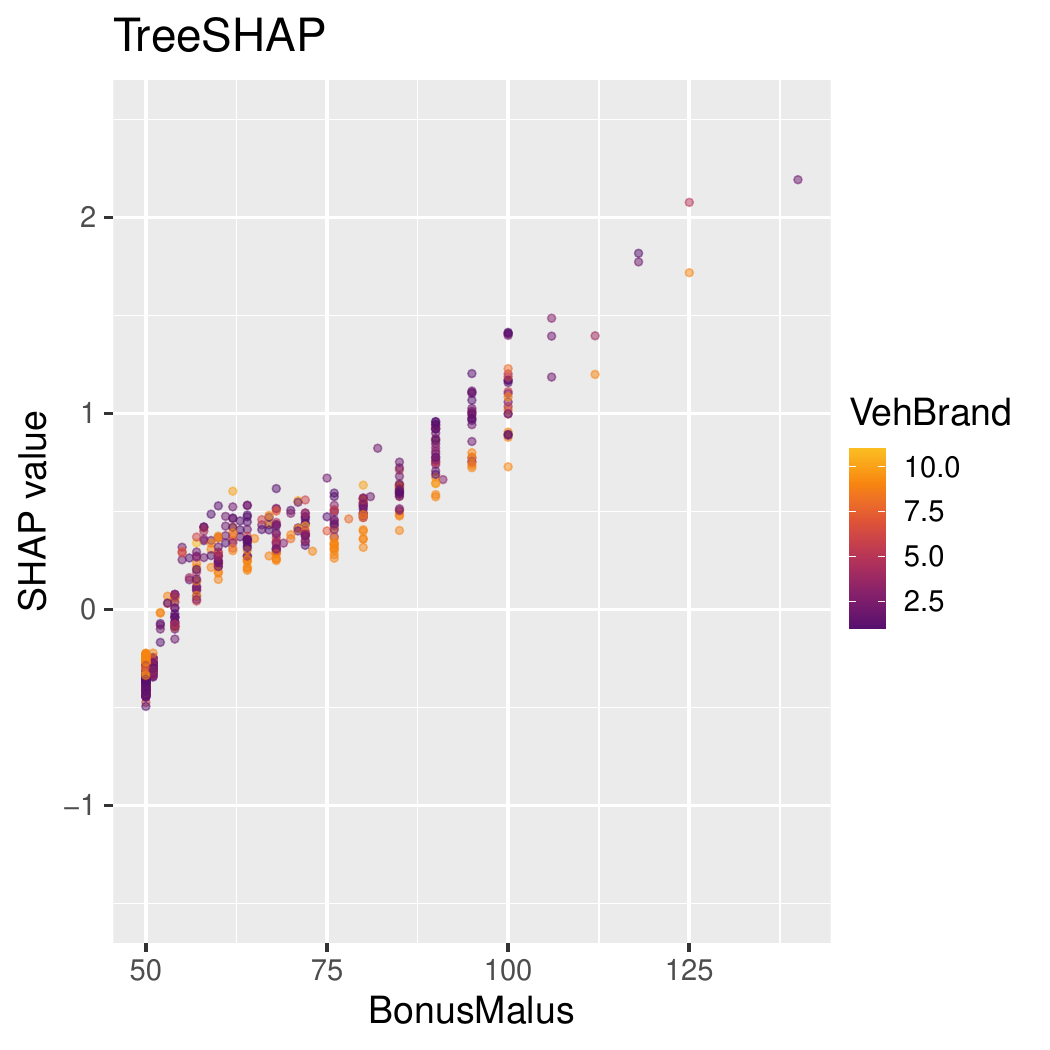}
\end{center}
\end{minipage}
\end{center}
\vspace{-.7cm}
\caption{Dependence plots of SHAP mean decompositions: (top) {\tt DrivAge}, (bottom) {\tt BonusMalus}, (lhs) conditional expectation
version \eqref{conditional expectation value}, (middle)
unconditional expectation version
\eqref{conditional expectation value independence}, (rhs) LightGBM surrogate model.}
\label{LightGBM}
\end{figure}

Figure \ref{LightGBM} gives a scatter plot of the two
feature components {\tt DrivAge} (top) and {\tt BonusMalus} (bottom)
on the $x$-axis vs.~their SHAP attributions on the $y$-axis for 
1000 randomly selected cases $\bx_i$;
remark that the two selected components
are dependent, see Figure \ref{colinearity plot} (lhs), and they
are expected to interact in the regression model. In Figure \ref{LightGBM} we show
the following SHAP mean attributions of 1000
randomly selected cases $\bx_i$: (lhs) conditional mean
\eqref{conditional expectation value} as value function, (rhs) unconditional mean \eqref{conditional expectation value independence} as value function,
and  (rhs)
TreeSHAP LightGBM surrogate model decomposition.
For the latter we use the {\sf R} package {\tt shapviz} \cite{shapviz},
which performs the decomposition on the log-scale, therefore, we also
choose the log-scale for the former two methods. The coloring
in Figure \ref{LightGBM} selects the feature component that shows
the highest interaction with the selected one, i.e., explains best the vertical scattering: (top) for feature {\tt DrivAge}
this is the variable {\tt BonusMalus}; (bottom) 
for feature {\tt BonusMalus} there
are different choices between the different SHAP methods.
The conditional mean version \eqref{conditional expectation value} selects {\tt DrivAge}, whereas the
unconditional mean \eqref{conditional expectation value independence} and TreeSHAP versions select {\tt VehBrand}; note 
that this categorical variable is treated differently in the network approach
(embedding layers) and in the LightGBM (exclusive feature bundling to
bundle sparse (one-hot encoded) features). From these graphs, it seems
that the unconditional mean version \eqref{conditional expectation value independence} and the TreeSHAP version using
a surrogate LightGBM provide rather similar results, and they are
different from the conditional mean version \eqref{conditional expectation value};
the plots use the same scale on the $y$-axis. Since the unconditional version
cannot cope with colinearity in feature components, we give preference
to the results in the left column of Figure \ref{LightGBM} using
the conditional mean version \eqref{conditional expectation value}.
In particular, this is justified by the discussion in Section \ref{SHAP for mean decompositions}, namely, for small values of {\tt DrivAge} we cannot have
a low {\tt BonusMalus} level, and any extrapolation to this part of the feature space is arbitrary (but it will impact the results of the unconditional version).

\subsection{SHAP for out-of-sample deviance loss attribution}
We have seen in Section \ref{Example: variable importance} that the {\tt anova} analysis depends on the order
of the inclusion of the feature components, i.e., we receive an additive loss decomposition which cannot be
considered to be fair, because if we change the order of inclusion
of the components their importance in the {\tt anova} analysis may
change. Instead of the {\tt anova} decomposition, we consider a Shapley 
deviance loss attribution in this section.
For this, we choose the value function of instance $\bx_i$ as
\begin{equation}\label{value loss decomposition}
{\cal C}~\mapsto~
  \nu_{(Y_i,\bx_i)}({\cal C}) := L \left(Y_i, 
    \mu_{\cal C}(\bx_i)\right)=
  L \Big(Y_i,\, \E \left[\left.\left. \mu(\bX) \right| \bX_{\cal C}=\bx_{i,\cal C} \right.\right]\Big),
\end{equation}
where $L$ is the Poisson deviance loss used, e.g., in \eqref{total increase},
and we add the specific choice of the observation $(Y_i,\bx_i)$ as a lower index
to the notation of the value function $\nu_{(Y_i,\bx_i)}$.
Note that we do not decompose the regression function $\mu(\bx)$ in this
section, but rather the resulting deviance loss $L(Y,\mu(\bx))$.

For ${\cal C}=\emptyset$ we receive the average loss of the null model
\begin{equation*}
\frac{1}{n}\sum_{i=1}^n L(Y_i, \mu_0) = \frac{1}{n}\sum_{i=1}^n \nu_{(Y_i,\bx_i)}(\emptyset),
\end{equation*}
and for ${\cal C}={\cal Q}$ we obtain the average loss of the full model
\begin{equation*}
\frac{1}{n}\sum_{i=1}^n L\left(Y_i, \mu(\bx_i)\right) = \frac{1}{n}\sum_{i=1}^n \nu_{(Y_i,\bx_i)}({\cal Q}),
\end{equation*}
this refers to lines (2) and (3) of Table \ref{hyperparameter 2}. Remark
that these quantities are empirical counterparts of the losses of the true
random tuple $(Y,\bX)$, given by for the null and the full model, respectively,
\begin{equation*}
\E \left[ L(Y,\mu_0) \right]=
\E \left[ \nu_{(Y,\bX)}(\emptyset) \right] \qquad \text{ and }
\qquad 
\E \left[ L(Y,\mu(\bX)) \right]=
\E \left[ \nu_{(Y,\bX)}({\cal Q}) \right].
\end{equation*}
Using the Shapley decomposition of Section \ref{Additive and fair mean decomposition}, we can attribute the
difference in these losses to the feature components $X_j$ of $\bX$.
In a first step, we therefore decompose the Poisson deviance loss 
$L(Y_i, \mu(\bx_i))$ for each 
observation $(Y_i,\bx_i)$ of the test sample ${\cal T}$
using the value function \eqref{value loss decomposition}. This provides
us for all observations $(Y_i,\bx_i)$ with an additive and 
fair decomposition giving the Shapley values
$(\phi_{j,(Y_i,\bx_i)})_{j=1}^q$ such that
\begin{equation*}
L\left(Y_i, \mu(\bx_i)\right)~=~
\nu_{(Y_i,\bx_i)}({\cal Q})~=~ L(Y_i, \mu_0) + \sum_{j=1}^q
\phi_{j,(Y_i,\bx_i)}.
\end{equation*}
In a second step, we average over these decompositions
to receive the average contribution (averaged over ${\cal T}$) of feature component $X_j$, $j\in {\cal Q}$, given by
\begin{equation}\label{SHAP loss decomposition}
\Phi_{j} = \frac{1}{n}\sum_{i=1}^n
\phi_{j,(Y_i,\bx_i)}.
\end{equation}
Since the Shapley decomposition 
is still computationally demanding, we consider 
\eqref{SHAP loss decomposition} for a random sub-sample of ${\cal T}$, otherwise one may use parallel computing.\footnote{To
  compute the Shapley decomposition of  the Poisson deviance loss for 1000 observations $(Y_i,\bx_i)$ on an ordinary laptop (based on a neural network
  ${\rm NN}_{\widehat{\bvartheta}}$) takes roughly 1 minute.}

\medskip

The following gives a pseudo-code for the SHAP deviance loss attribution.
  
\medskip  

\hrulefill

\begin{itemize}
\item[(0)] Select at random a fixed number $m\le 2^q-2$ of non-empty subsets 
${\cal C} \subset {\cal Q}$, and calculate for this random selection the
matrix, see \eqref{efficient Shapley}, 
\begin{equation*}
A =\left(Z^\top W  Z\right)^{-1} Z^\top W ~\in ~\R^{q \times (m+1)},
\end{equation*}
where we additionally add the case ${\cal C}={\cal Q}$ with
a large Shapley kernel weight.
\item[(1)] Select at random a fixed number $n$ of cases $i$, and calculate
for each case $i$ and each selected subset ${\cal C}$ from item (0) the
individual deviance loss differences, see \eqref{value loss decomposition} and
\eqref{masked x},
\begin{equation}\label{efficient SHAP}
  \widehat{\nu}^0_{(Y_i,\bx_i)}({\cal C}) ~=~ L \left(Y_i, 
    {\rm NN}_{\widehat{\bvartheta}}(\bx^{(\bm)}_{i,{\cal C}})\right)
    - L(Y_i,\mu_0).
\end{equation}
This provides us with vectors 
$\widehat{\bnu}_i=(\widehat{\nu}^0_{(Y_i,\bx_i)}({\cal C}))_{{\cal C}}
\in \R^{m+1}$,
considering all the subsets ${\cal C}\subseteq {\cal Q}$ selected in item (0).
\item[(2)] Compute the approximate individual Shapley 
deviance loss decompositions 
of all selected cases $i$
\begin{equation*}
(\widehat{\phi}_{j,(Y_i,\bx_i)})_{j=1}^q =   A \widehat{\bnu}_i~\in ~\R^q.
\end{equation*}
\item[(3)] Return the estimated average
attributions $\widehat{\Phi}_{j} = \frac{1}{n}\sum_{i=1}^n
\widehat{\phi}_{j,(Y_i,\bx_i)}$.
\end{itemize}

\hrulefill  
  
\medskip  
  
We give a few remarks. Matrix $A$ in item (0) only considers the
rows and columns of the Shapley kernel weight matrix $W$ and
the design matrix $Z$ that have been chosen by the random
selections ${\cal C}\subseteq {\cal Q}$, we also refer to \eqref{efficient Shapley}.
This is an approximation that reduces the computational complexity for large
$q$. Item (1) uses the network approximation for the calculation
of the conditional expectations \eqref{conditional expectation}. This
step precisely reflects the efficiency gain of our approach because it
requires only {\it one single} fitted neural network ${\rm NN}_{\widehat{\bvartheta}}$ to calculate the conditional expectations for all considered cases $i$ and
all selected subsets ${\cal C}$, see \eqref{efficient SHAP}. Item (2) are simple matrix multiplications
that always rely on the same matrix $A$.

\medskip

\begin{figure}[htb!]
  \begin{center}
\begin{minipage}[t]{0.48\textwidth}
\begin{center}
\includegraphics[width=\textwidth]{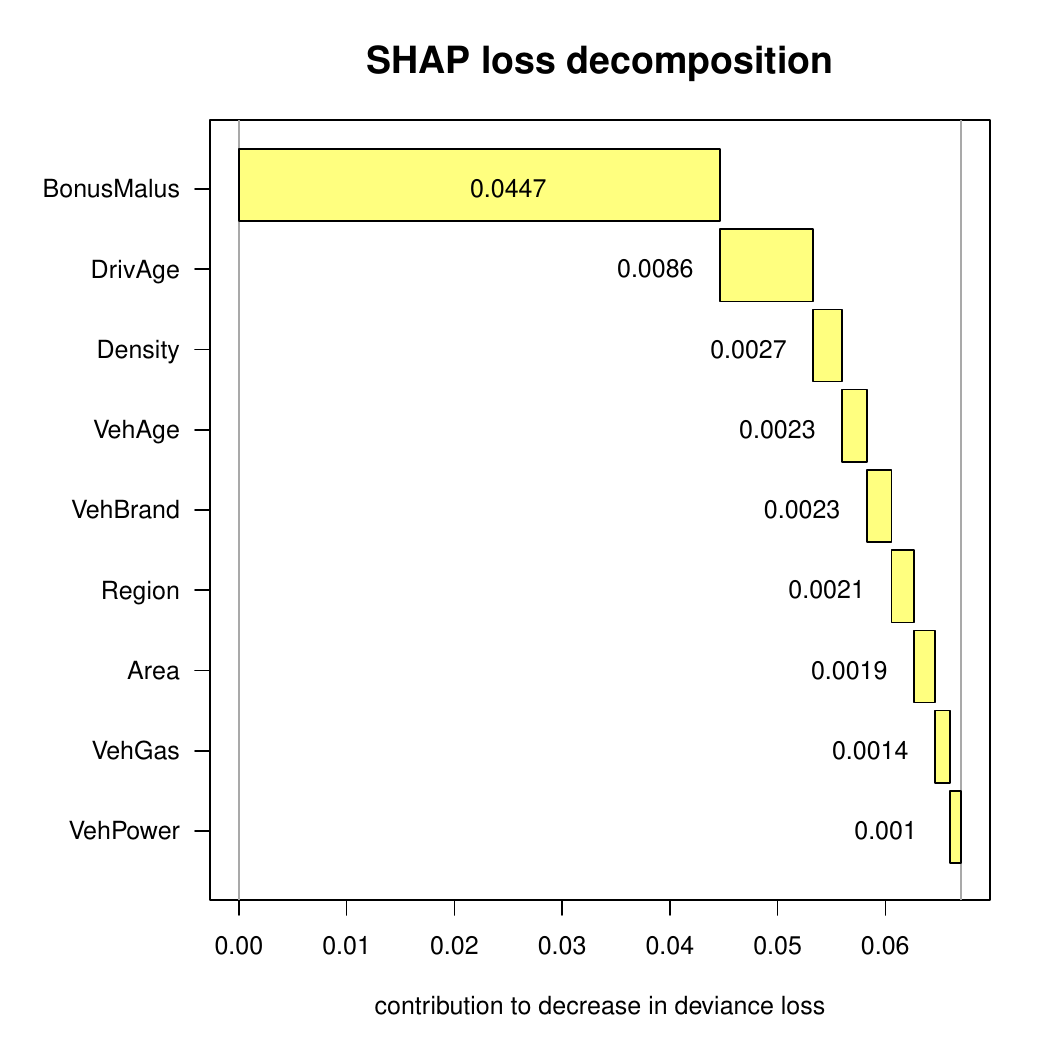}
\end{center}
\end{minipage}
\end{center}
\vspace{-.7cm}
\caption{SHAP Poisson deviance loss decomposition
$({\tt SHAP\_anova}_j)_{j \in {\cal Q}}$.}
\label{SHAPANOVA}
\end{figure}

Figure \ref{SHAPANOVA} shows the resulting (relative) SHAP Poisson deviance loss decomposition for, $j\in {\cal Q}$,
\begin{equation*}
{\tt SHAP\_anova}_j~=~-
\frac{\Phi_{j}}{\frac{1}{n}\sum_{i=1}^n L\left(Y_i, \mu(\bx_i)\right)},
\end{equation*}
compare to \eqref{anova formula};
the $y$-scale
is in the order of the magnitudes of the decreases. This figure should be compared
to the {\tt anova} analyses of Figure \ref{ANOVA}. In contrast to these latter graphs,
Figure \ref{SHAPANOVA} provides a variable importance ranking that is fair (in the Shapley sense and
for the chosen value functions \eqref{value loss decomposition}), and it
does not depend on the order of the inclusion of the feature components. E.g., we observe that the magnitudes of the contributions
of {\tt BonusMalus} and {\tt DrivAge} are somewhere in between the values in
Figure \ref{ANOVA}, where these two variables are included in different orders in this latter figure. From Figure \ref{SHAPANOVA}
we mainly question the importance of {\tt VehPower}, and we could rerun the models without this variable.

Another interesting observation is the importance of {\tt Density} and {\tt Area} which are highly colinear,
see Figure \ref{colinearity plot} (rhs).
Both variables receive a similar magnitude of importance, the more granular {\tt Density} variable being slightly more important. We interpret
these SHAP results in case of colinear variables as follows. These two
variables share a (common) importance because they may
equally contribute to the decrease in loss, i.e., we have 
(almost) equally behaved players in this cooperative game.
Of course, this also means that the importance of a variable is
diminished if we add a second colinear one to the model, and, in fact, we should
work with the smaller model in that case. This is not in contradiction to the examples in
Sundararajan--Najmi \cite{Sundar}, it just says that colinearity needs to be assessed carefully before
regression modeling, because the regression model may not detect this (and
we should try to work in a parsimonious model already in the first place).
A way of exploring colinearity is the {\tt anova} graph of Figure \ref{ANOVA} because changing
the order of inclusion will also change the magnitudes of contribution, as can be seen from that
figure for the variables {\tt Density} and {\tt Area}.

\section{Conclusion}
\label{Conclusions}
Starting from a regression function $\mu(\bX)$ that is based
on tabular input data $\bX$, we have proposed a neural network
surrogate model that can calculate the conditional expectations
of $\mu(\bX)$ by conditioning on any subset of components
of the tabular input data $\bX$. These conditional expectations are useful
in different contexts. We present an {\tt anova} and a {\tt drop1}
variable importance analysis, respectively. These analyses are similar to their generalized
linear model (GLM) counterparts, except that we do not require to 
have nested models, here, but we start from a bigger model and
calculate the smaller model. Our second example modifies the partial dependence
plot (PDP) by correcting for the deficiency that PDPs cannot cope with dependence
structures in the features $\bX$. Our proposal, the marginal conditional expectation plot (MCEP),
correctly considers these dependence structures and it provides convincing explainability
results that reflect the empirical observations.
Our third example concerns the {\bf SH}apley {\bf A}dditive ex{\bf P}lanation (SHAP). We show that the neural network
surrogate model for conditional expectations allows us to efficiently
calculate the conditional SHAP decompositions both, for the mean
but also for the decrease in deviance losses of the full model against
the null model. The latter provides us with an interesting method
for variable importance.

\bigskip

{\bf Acknowledgment.} We kindly thank Michael Mayer for his methodological support
and for supporting us in improving the figures.

\bigskip

{\small 
\renewcommand{\baselinestretch}{.51}

}

\end{document}